\documentclass{article}

\usepackage{iclr2026_conference,times,lila}
\preprint
\usepackage{amsmath}
\usepackage{amssymb}
\usepackage{booktabs}
\usepackage{graphicx}
\usepackage{microtype}
\usepackage{url}
\usepackage{float}
\usepackage{algorithm}
\usepackage{algpseudocode}
\usepackage{caption}
\usepackage{subcaption}
\algrenewcommand\algorithmicrequire{\textbf{Input:}}
\usepackage{adjustbox}
\usepackage{tabularx}
\usepackage{array}
\usepackage{makecell}
\usepackage{cuted}
\usepackage{wrapfig}
\usepackage{hyperref} 
\usepackage{etoolbox}
\AtBeginEnvironment{abstract}{%
  \setlength{\leftmargini}{0pt}%
}

\newlength{\featuremapheight}
\newlength{\heroheight}
\newcommand{\coderelease}{\href{https://github.com/fl97inc/are-ufos-real}{\textcolor{blue}{\nolinkurl{github.com/fl97inc/are-ufos-real}}}}
\newcommand{\blfootnote}[1]{{\let\thefootnote\relax\footnotetext{#1}}}

\makeatletter
\AtBeginDocument{%
  \def\@maketitle{\thispagestyle{lilafirstpage}\vbox{\hsize\textwidth
    \team@authorsetup
    {\LARGE\bfseries\@title\par}%
    \vskip 0.14in
    \team@authors
    \vskip 0.3in minus 0.1in}}%
  \renewenvironment{abstract}{%
    \vskip.12in
    {\color{TeamCopperMid}\hrule height 2.5pt width 0.5in}%
    \vskip 0.06in
    {\large\color{black}\textbf{Abstract}\par}%
    \vspace{-1ex}\begin{quote}}{\par\end{quote}\vskip 1ex}%
}

\title{Are UFOs Real?\\Introducing Unified Factored Organizers\vspace{-0.30in}}

\author{Thomas X. Sheppard, Alon Albalak,  Joel Lehman, Kenneth O. Stanley \\
Lila Sciences \\
\texttt{tsheppard@lila.ai}
}

\begin{document}

\vspace*{-0.5in}
\maketitle
\blfootnote{Code for the grokking experiments is available at \coderelease.}
\vspace{-1.75em}

\begin{figure}[H]
    \centering
    \includegraphics[height=\heroheight,width=\linewidth,keepaspectratio]{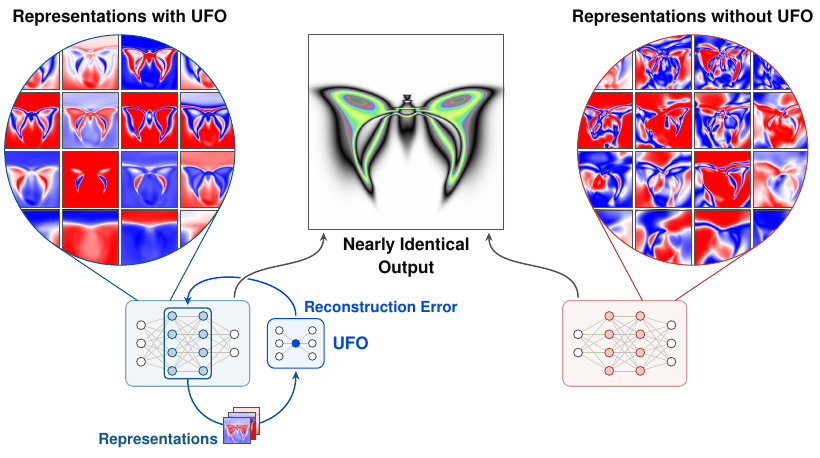}
    \captionsetup{skip=4pt}
    \caption{\textbf{Representations from networks trained with and without UFO regularization.}
    Both depicted networks are trained to output a butterfly, but the blue network is also trained with a
    \emph{unified factored organizer} (UFO) that encourages regular representations. Each square
    within the two large circles shows the activation pattern of one hidden node. Although the two networks
    produce nearly identical outputs, the blue (UFO-regularized) network's representations better respect the underlying regularities of the butterfly. 
    }
    \label{fig:picbreeder-hero}
\end{figure}
\vspace{-0.25em}

\begin{abstract}

Implicit in much research in deep learning is the assumption that if a neural network achieves low loss, then it must have learned a good representation of its training data. Recent work by \citet{kumar2025questioningrepresentationaloptimismdeep} questions this assumption, arguing that neural networks trained with conventional gradient descent achieve low loss despite learning \emph{fractured entangled representations} (FER), leaving related concepts disconnected and unrelated concepts entangled inappropriately. \citet{kumar2025questioningrepresentationaloptimismdeep} further show that dramatically better representations are possible: a network evolved through an open-ended, human-in-the-loop process learned \emph{unified factored representations} (UFR) that capture the regularities of the domain far better than a counterpart with nearly identical outputs trained with conventional gradient descent. Inspired by UFR, this paper introduces \emph{unified factored organizers} (UFOs), neural networks that are specially trained to model regularities in the representations of other neural networks. A neural network can then be trained both to achieve low loss on its task objective and to align its representations with the model learned by the UFO. Neural networks trained with UFOs are shown to learn qualitatively superior representations in an image-regression domain where representations are easily interpretable. To provide quantitative evidence for the impact of UFOs on representation learning, a transformer is trained with and without UFO regularization to grok a modular arithmetic problem set. When UFOs are combined with weight decay, UFOs can reduce the number of transformer training steps needed to grok by more than $10\times$ relative to a baseline regularized with weight decay alone. Without weight decay, UFOs can reduce the number of transformer training steps needed to grok by more than $100\times$ relative to an unregularized baseline.  If UFOs are real, they suggest a new paradigm where loss functions themselves are trained to learn an effective model of good representation.

\end{abstract}

\clearpage

\section{Introduction}
\label{sec:introduction}

Neural networks trained with conventional gradient descent can reliably achieve low loss on their training objective, yet often still exhibit surprising deficiencies. Neural networks can have dangerous failure modes when evaluated beyond their training data distribution \citep{koh2021wildsbenchmarkinthewilddistribution, wei2023jailbroken}, catastrophically forget what they previously learned when tasked with integrating new data \citep{MCCLOSKEY1989109, kemker2018aaai-measuring, yuocatastrophicforgetting}, or can be severely miscalibrated, giving confidently wrong outputs \citep{guo2017calibrationmodernneuralnetworks, xiong2024can, damani2026beyond}. Recent work by \citet{kumar2025questioningrepresentationaloptimismdeep} argues that these surprising deficiencies result from the \emph{fractured entangled representations} (FER) that models tend to learn through  conventional gradient descent, despite achieving low loss. FER means that the underlying representations of trained networks leave related concepts disconnected, while unrelated concepts are entangled inappropriately. \citet{kumar2025questioningrepresentationaloptimismdeep} suggest that neural networks should learn \emph{unified factored representations} (UFR) that respect the underlying regularities of the training data.  Until now, no general method has existed to encourage UFR.

Representational quality is clearly important, but FER and UFR remain challenging to study. Although mechanistic interpretability has made some progress in describing specific features or circuits, neural network representations are still largely opaque \citep{10136140, sharkey2025openproblemsmechanisticinterpretability}. Even if neural network representations are easier to understand than originally thought, as recently suggested by \citet{cammarata2026clean}, mechanistic interpretability today focuses to a large extent on \emph{describing}, rather than \emph{evaluating}, representations. This presumption that the field need only develop a descriptive theory of representation reflects widespread representational optimism, the view that networks that achieve low loss invariably learn good representations. Representational optimism may in turn cause the field to overlook the potential importance of evaluating representations to distinguish between FER and UFR. \citet{kumar2025questioningrepresentationaloptimismdeep} argue that low loss is not a reliable indicator of representational quality by contrasting example networks with identical outputs but representations of radically different quality. 

Given that a mechanistic understanding of FER and UFR is still undeveloped and that loss alone cannot adjudicate representational quality, rare example networks that are known to have virtuous representations or have uniquely interpretable representations are particularly valuable to the study of FER and UFR. For this purpose, networks evolved through Picbreeder \citep{10.1145/1357054.1357328, Secretan2011PicbreederAC} stand out as perhaps the best such example.  Picbreeder was an experimental website where users could interactively evolve pictures, and each picture was output by a corresponding neural network. Interestingly, networks evolved in Picbreeder are adaptable, evolvable, and amenable to continual learning, providing evidence of UFR \citep{10.1162/artl_a_00263, kumar2025questioningrepresentationaloptimismdeep, lehman2026unified}. Furthermore, conveniently, their representations can be explicitly visualized as images. Motivated by the simplicity and regularity of Picbreeder hidden node activation patterns, this paper builds on the hypothesis that such regularity is essential to good representation. 

To further explore the implications of this hypothesis, the main idea is to learn an auxiliary loss function that models and encourages regular representations. This auxiliary loss function is an autoencoder \citep{rumelhart:errorpropnonote, hinton2006reducing} that is trained to reconstruct the representations of a primary neural network. The primary neural network is trained both to minimize a task objective and \emph{to make its representations compressible by the autoencoder}. Because regular representations are more likely to be compressible in this way, the primary neural network is encouraged not only to achieve low loss on its primary task objective, but also to achieve such low loss with regular, compressible representations. The autoencoder-based auxiliary loss function is intended to lead to UFR as seen in Picbreeder networks, so we refer to the autoencoder in this context as a \emph{unified factored organizer} (UFO). 

\begin{figure*}[!htbp]
    \centering
\includegraphics{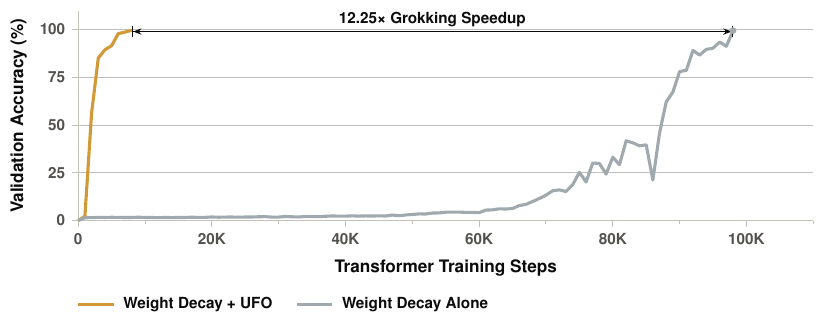}
    \caption{\textbf{Median grokking speedup caused by the fastest UFO treatment combined with weight decay when compared to a transformer trained with weight decay alone.} Two transformers with matching hyperparameters are trained on division modulo $97$, one with UFO regularization and weight decay (gold) and one with weight decay alone (gray). The transformer trained with UFO regularization groks (i.e.\ achieves $\geq 99\%$ validation accuracy) in $\sim12.25\times$ fewer transformer training steps than the transformer trained with weight decay alone. This pair of runs is highlighted because each run grokked at its treatment's median grok step across eight seeds, showing the median grokking speedup induced by the fastest UFO treatment.} 
    \label{fig:median-speedup-combined-with-weight-decay}
\end{figure*}

In addition to showing that UFOs can encourage qualitatively superior representations in neural networks trained to reproduce Picbreeder images, experiments in this paper also demonstrate that UFOs can dramatically reduce the number of training steps a transformer \citep{vaswani2023attentionneed} needs to grok modular division problems, establishing concrete and significant implications for learning.  Grokking, originally introduced by \citet{power2022grokkinggeneralizationoverfittingsmall}, refers to the phenomenon of delayed generalization to the validation data after a neural network has already fit its training data. Grokking is often demonstrated on modular arithmetic problem sets. As an arithmetic operation $\circ$ and a prime $p$ determine a fixed, finite set of modular arithmetic problems, when a network ``groks'' and generalizes to the validation data, it has learned to solve \emph{all} problems in the domain. As such, grokking modular arithmetic problems gives strong evidence that a neural network has learned good representations of its training domain. When UFOs are combined with weight decay, they can reduce the number of transformer training steps needed to grok by more than $10\times$ relative to a baseline regularized with weight decay alone, and without weight decay UFOs can reduce the number of transformer training steps needed to grok by more than $100\times$ relative to an unregularized baseline. The median grokking speedup for the fastest UFO treatment relative to a baseline trained with weight decay alone is depicted in Figure~\ref{fig:median-speedup-combined-with-weight-decay}.

The important question raised by these results that we hope will inspire further research is what the downstream implications are for such UFO-induced representations.  The qualitative improvements in the underlying representations of networks trained to reproduce Picbreeder images plus the dramatic acceleration of grokking hint at implications for generalization, continual learning, and creativity, as argued by \citet{kumar2025questioningrepresentationaloptimismdeep} for UFR. In effect, while Picbreeder showed that UFR exists, until now there has been no straightforward method to \emph{induce} UFR.  
While \citet{kumar2025questioningrepresentationaloptimismdeep} hinted that an explicitly open-ended search process like in Picbreeder might be necessary to arrive at UFR, the results herein suggest that it may actually be possible to encourage gradient descent to walk a more representation-friendly trajectory through clever regularization.
With such a method now at hand with the UFOs introduced in this paper, it is possible to explore UFR's implications across many domains and architectures, including for LLMs.
  
\section{Background}
\label{sec:background}

This section first reviews Picbreeder and its connection to FER and UFR. It then turns to grokking and the representational changes that underlie it. Related regularization methods are discussed in Section~\ref{sec:related-work} after UFOs have been explained in greater detail. 

\subsection{Picbreeder, FER, \& UFR}
\label{sec:background-picbreeder}

Picbreeder \citep{10.1145/1357054.1357328, Secretan2011PicbreederAC} was an experimental website where users could select pictures to breed and produce offspring pictures, which are then presented for a subsequent round of selection and breeding. By repeatedly selecting pictures for breeding, users can evolve abstract geometric patterns into recognizable pictures of butterflies, skulls, cars, and more. Each picture in Picbreeder is rendered by a separate compositional pattern producing network (CPPN), a kind of neural network inspired by evolutionary developmental biology \citep{Stanley2007CompositionalPP}. CPPNs whose corresponding picture is selected for breeding are then evolved to form the next generation of pictures based on a version of the NeuroEvolution of Augmenting Topologies (NEAT) algorithm \citep{neat2002}. 

CPPNs render images by mapping each pixel coordinate to a color. Specifically, an input tuple consisting of pixel coordinates $x$, $y$, distance from the origin $d = \sqrt{x^2 + y^2}$, and a scalar bias is mapped to a hue, saturation, and value output tuple. As each pixel in the image corresponds to an input-output pair for the CPPN, the CPPN-rendered image is a visualization of its behavior over \emph{all} inputs. As such, a CPPN image can be regarded as a metaphor for the total possible output behavior of a much larger neural network.

\begin{figure*}[!htbp]
    \centering
\includegraphics{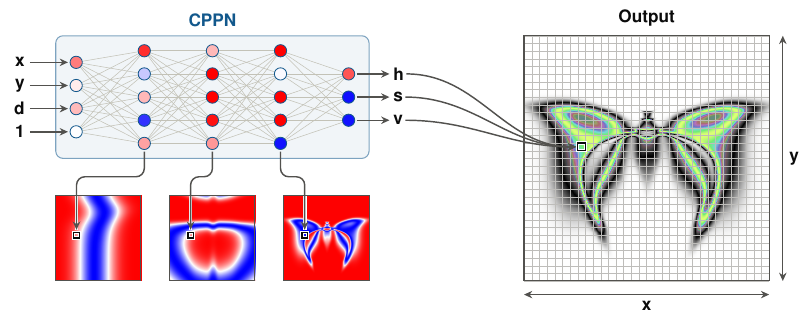}
    \caption{\textbf{Visualizing CPPN outputs and hidden node activations as images}. This figure visualizes a CPPN that evolved in Picbreeder to output a butterfly. An input tuple of pixel coordinates $x$, $y$, a distance from the origin $d = \sqrt{x^2 + y^2}$, and a scalar bias is passed into the CPPN. The CPPN then outputs a hue, saturation, value output tuple. This output tuple corresponds to the output image's pixel color at those pixel coordinates $x$, $y$. The CPPN's entire output behavior can then be visualized as an image by showing the output pixel color for all pixel-coordinate inputs. Likewise, the activations of any hidden node can be visualized as an image by recording the hidden node's activations over all pixel-coordinate inputs. The activation map images of three hidden nodes from the Picbreeder butterfly network are shown on the bottom left. For these hidden nodes, a line connects a hidden node's activation for input pixel coordinates $x$, $y$ to its corresponding position on the activation map image. This figure is inspired by visualizations from \citet{kumar2025questioningrepresentationaloptimismdeep}. 
    }
    \label{fig:cppn-image-visualizations}
\end{figure*}

Conveniently for interpretability, just as the output of the CPPN can be visualized as an image, the activations of \emph{any hidden node} can be visualized as an image as well, simply by recording that hidden node's activation intensity for each pixel-coordinate input. Figure~\ref{fig:cppn-image-visualizations} illustrates how a CPPN's output and hidden node activations can be visualized as images. While it is intractable to visualize how an LLM responds across the vast space of possible inputs, CPPNs offer a microcosm for studying how internal representations relate to output behavior.

\begin{figure*}[!tp]
    \centering
    \includegraphics[
        width=\textwidth,
        height=\featuremapheight,
        keepaspectratio
    ]{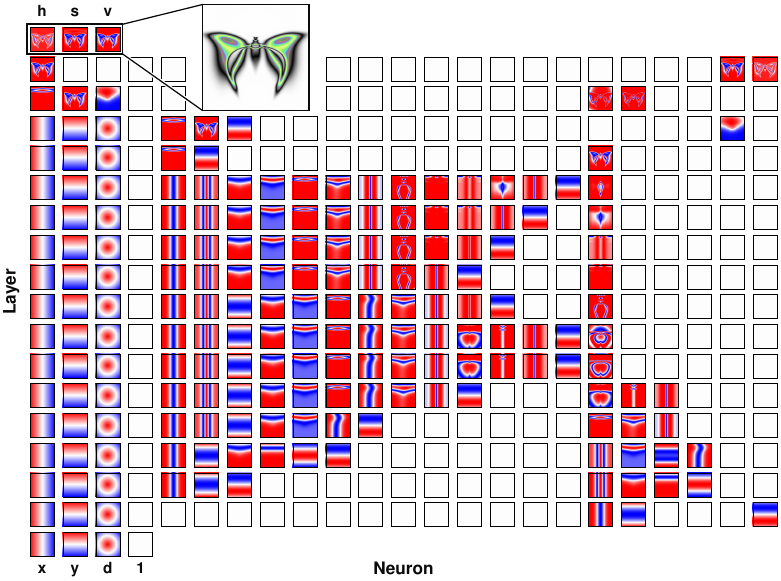}  
    \caption{\textbf{Representations of a CPPN that evolved in Picbreeder to output a butterfly.} This visualization approach was inspired by \citet{kumar2025questioningrepresentationaloptimismdeep}, although individual hidden node activation maps may differ in color scheme from the original presented in that work. Each row in the figure corresponds to a hidden layer in the Picbreeder CPPN, and each square corresponds to an individual hidden node. As the CPPN maps pixel coordinates to pixel colors, the butterfly image is the network's output over all pixel-coordinate inputs. Likewise, the image for each hidden node captures that hidden node's activations over all pixel-coordinate inputs. Each hidden node's activations are rescaled independently, and red is used for the lowest activation, blue for the highest activation, and white for the midpoint of the activation range. Constant or unused nodes are left white. As can be seen, hidden node activations respect the symmetry and structure of the butterfly, which led \citet{kumar2025questioningrepresentationaloptimismdeep} to conclude that this Picbreeder network has learned \emph{unified factored representations} (UFR) of the butterfly. This representation visualization is over an MLP architecture that is computationally equivalent to the original Picbreeder-evolved CPPN architecture. This MLP version is constructed through a layerization process. 
    Layerization for Picbreeder networks is described in \citet{kumar2025questioningrepresentationaloptimismdeep}; a similar process was first described in chapter 3.2 of \citet{budenbender2021revisiting}. 
}
    \label{fig:picbreeder-butterfly}
\end{figure*}

CPPNs that map pixel coordinates to colors afford the rare opportunity for intuitive visualizations of the entirety of a network's internal representation across its dataset. Figure \ref{fig:picbreeder-butterfly} visualizes the activations of each hidden node in a CPPN that evolved in Picbreeder to output a butterfly. The activation patterns respect the underlying symmetry of the butterfly --- activation maps of individual hidden nodes are highly symmetric, and the patterns within and across layers are regular and simple. For this reason (and based on weight sweep experiments within the weight space of such networks), \citet{kumar2025questioningrepresentationaloptimismdeep} conclude that the Picbreeder network has learned \emph{unified factored representations} (UFR) of the butterfly. 

Because this paper builds on \citet{kumar2025questioningrepresentationaloptimismdeep}, it is important to review their findings. Beyond establishing the Picbreeder butterfly (among other Picbreeder networks) as having UFR, \citet{kumar2025questioningrepresentationaloptimismdeep} also compare these UFR to the representations of networks trained with conventional gradient descent. A network trained with conventional gradient descent to reproduce the Picbreeder butterfly image shows far less representational regularity than the original Picbreeder network, despite achieving low loss and outputting a nearly identical image \citep{kumar2025questioningrepresentationaloptimismdeep}. \citet{kumar2025questioningrepresentationaloptimismdeep} propose that this conventional gradient descent network has learned \emph{fractured entangled representations} (FER) of the butterfly --- activation patterns are asymmetric, high-frequency, and do not respect the salient structures of the butterfly. This experiment is recreated in Section~\ref{sec:picbreeder-experiments}, and the representations of a network trained with conventional gradient descent to output the butterfly are shown in Figure~\ref{fig:sgd-butterfly}. In addition, entirely new kinds of improved representations caused by UFOs are shown in Figures \ref{fig:intracomponent-ufo-butterfly} and \ref{fig:intercomponent-ufo-butterfly}. In short, the UFO method introduced here aims to enable gradient descent to achieve improved internal representations (like those of the Picbreeder networks).

In this way, this paper can be seen as contributing to a growing body of literature on better understanding Picbreeder and using image-rendering CPPNs to study representation. For example, \citet{earle2026picbreedervlm} recreate the original Picbreeder site and let AI agents evolve images to see if they can match the creative outputs of the original, human-driven Picbreeder experiment. Other works aim to explain and improve representations in image-rendering CPPNs by applying image analysis techniques such as total variation to network activations \citep{xu:hthesis26} or characterizing representational quality as a function of network depth and sparsity \citep{do2026compositionalityemergesnarrowdepthconnectivity}. Overall, Picbreeder images rendered by CPPNs provide a convenient, intuitive context for understanding and visualizing the impact of different training approaches on representation, which is why it is chosen as an experimental domain in this paper.

\subsection{Grokking}
\label{sec:background-grokking}

\begin{figure*}[!htbp]
    \centering
    \includegraphics[
        width=\textwidth,
        height=0.3\textheight,
        keepaspectratio
    ]{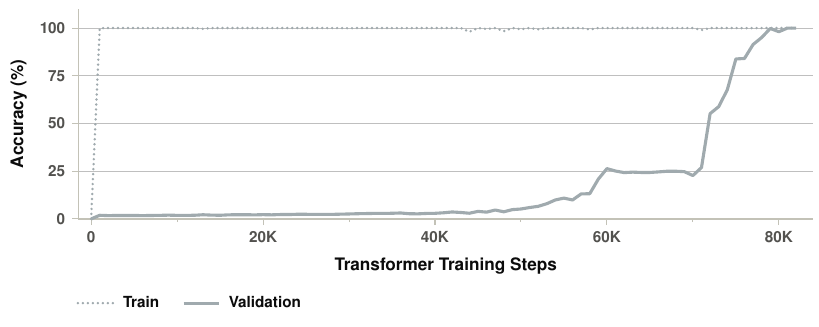}
    \caption{\textbf{Train and validation accuracy curves that demonstrate grokking with a transformer trained on division modulo 97.} To demonstrate typical grokking behavior, a transformer is trained on division modulo $97$ with weight decay and without UFO regularization until it groks (i.e.\ reaches $\geq 99 \%$ validation accuracy). The transformer fits the training data and achieves $\geq 99 \%$ training accuracy in less than $1$,$000$ steps. After $\sim79$,$000$ steps, the transformer generalizes to the validation data and groks. For tens of thousands of steps after the transformer has fit the training data, validation accuracy stays near $0\%$. Although the network's delayed generalization appears sudden, the representations of the transformer when it has grokked emerge gradually, as reported by \citet{nanda2023progressmeasuresgrokkingmechanistic} and depicted in Figure~\ref{fig:trajectory-map}. This transformer is trained with the same settings as the weight decay control runs in Section~\ref{sec:grokking-experiments}.
    }
    \label{fig:normal-grokking-curve}
\end{figure*}

Grokking, the phenomenon of delayed generalization after initially fitting the training data, was discovered by \citet{power2022grokkinggeneralizationoverfittingsmall} when training small transformers on algorithmically generated datasets. Grokking can require training long after the model achieves  $100\%$ accuracy on the training data. Based on validation accuracy alone, grokking seems to occur suddenly because the validation accuracy begins to rise rapidly after being near zero for many steps. Figure~\ref{fig:normal-grokking-curve} shows train and validation accuracy curves that illustrate the grokking phenomenon.   

\begin{figure*}[!t]
    \centering
    \includegraphics{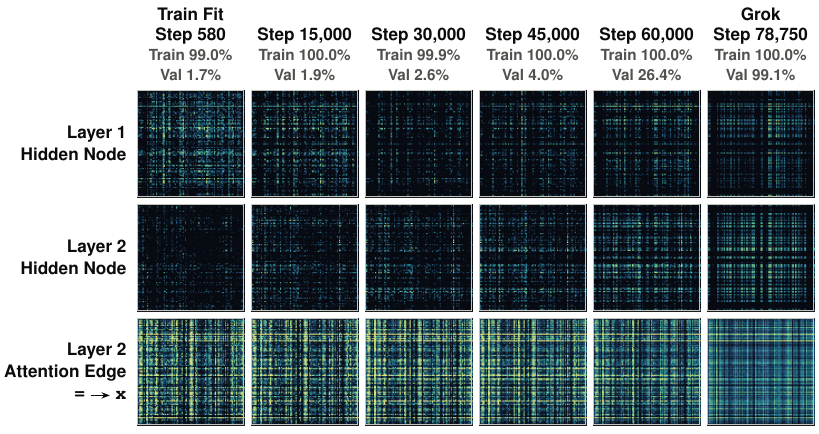}
    \caption{\textbf{Gradually emerging representations of a transformer trained on division modulo 97}. Response maps are shown for select components from the same $2$-layer transformer training run shown in Figure~\ref{fig:normal-grokking-curve}, which uses weight decay without UFO regularization and is trained with the same settings as the weight decay control runs in Section~\ref{sec:grokking-experiments}. The transformer is trained to predict the correct next token \texttt{z} for an equation of the form $x / y = z\pmod{97}$. For each square in the above graphic, the $x$ and $y$ axes span from $0$ to $97$, and each $(x,y)$ position captures a component's response for the operands $x$ and $y$.  Analogous to the hidden node activation maps in Figure~\ref{fig:picbreeder-butterfly}, each square captures a component's response over all operand pairs $(x,y)$, thus showing a component's response for all inputs in the domain. The first two rows show hidden nodes from the first and second transformer layers respectively, and the intensity at each $(x,y)$ position records the hidden node's activation for that $(x,y)$ input. The third row shows an attention edge in the second layer, where an \emph{attention edge} describes the query-key connection between two fixed sequence positions in a given attention head. Here, the edge connects the \texttt{=} and \texttt{x} tokens, and the intensity at each $(x,y)$ position records the corresponding attention weight for that $(x,y)$ input. Each component response map is scaled independently; dark colors indicate a low response and light colors a high response. Each column denotes a different training step, and accordingly each row shows how the responses of a component evolve over training. Although grokking appears sudden in Figure~\ref{fig:normal-grokking-curve}, the highly regular response patterns of each component at the grok step are apparent in fuzzy and partial ways earlier in training, consistent with \citet{nanda2023progressmeasuresgrokkingmechanistic}. Because these representational regularities are refined gradually, the UFO network can potentially detect and amplify these regularities prior to grokking.}
    \label{fig:trajectory-map}
\end{figure*}

Although grokking may appear sudden, \citet{nanda2023progressmeasuresgrokkingmechanistic} analyze a small transformer that is trained with weight decay to grok modular addition and reveal that the highly regular representations underlying grokking emerge gradually over training. Consistent with this analysis, Figure~\ref{fig:trajectory-map} shows regular structure incrementally taking shape in activations and attention weights from the same transformer training run depicted in Figure~\ref{fig:normal-grokking-curve}. This observation, that the regular activation and attention patterns that lead to grokking are gradually refined during training, helps motivate UFOs, which are intended to model representational regularities as they emerge throughout training and accelerate their formation when they are in a nascent stage. 

Much prior research has emphasized the importance of weight decay for accelerating grokking \citep{power2022grokkinggeneralizationoverfittingsmall, nanda2023progressmeasuresgrokkingmechanistic, liu2023omnigrokgrokkingalgorithmicdata}. Although weight decay is the best understood method for accelerating grokking, grokking can occur without it \citep{thilak2022slingshotmechanismempiricalstudy, gromov2023grokkingmodulararithmetic}, and even when the norm of the weights is increasing (which is what weight decay counteracts; \citealp{kumar2024grokkingtransitionlazyrich}). Highlighting that UFOs are an independent method for accelerating grokking, later results show that UFOs are effective even without weight decay. 

UFOs may encourage representations to lie near a low-dimensional activation manifold, so they relate to previous research connecting such low-dimensional structure to grokking and generalization. The intrinsic or effective dimensionality of hidden layer activations, especially at the deepest hidden layer, drops sharply prior to or around the onset of grokking \citep{brown2022relatingregularizationgeneralizationintrinsic, kazanskii2026tamegrokkingrepresentationgeometry}. Beyond the context of grokking, hidden layer activations of trained networks lie near low-dimensional, curved manifolds with an intrinsic dimension that can be orders of magnitude smaller than the layer width \citep{recanatesi2019dimensionalitycompressionexpansiondeep, ansuini2019intrinsicdimensiondatarepresentations}, and the last hidden layer's intrinsic dimensionality can strongly predict test accuracy \citep{ansuini2019intrinsicdimensiondatarepresentations}.

In addition to weight decay and the UFOs presented here, other work introduces techniques to accelerate grokking \citep{lee2024grokfastacceleratedgrokkingamplifying, xu2025letgrokyouaccelerating, kazanskii2026tamegrokkingrepresentationgeometry}. In contrast to these methods, UFOs are motivated by a new hypothesis about grokking and representation learning: the representations of a network that generalizes should be regular and therefore learnable by a separate, capacity-limited network. Conceptually, if a network has arrived at a memorized solution and there are no regularities between the representations used for one training example and another, then its representations should be less learnable by a separate, capacity-limited network. 

\section{Method: Unified Factored Organizers (UFOs)}
\label{sec:method}

\begin{figure*}[!htbp]
    \centering
\includegraphics{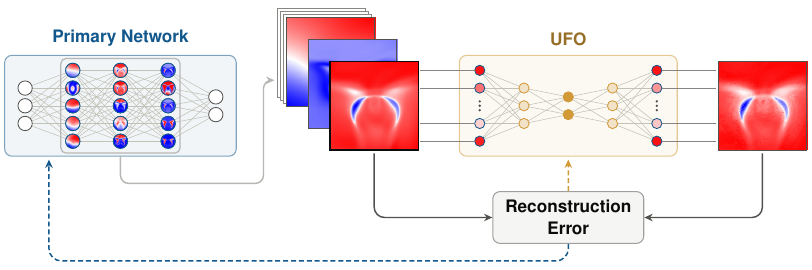}
    \caption{\textbf{The UFO method}. The activations of each hidden node for a training batch are extracted, normalized, and inputted into the UFO autoencoder. In networks trained to output Picbreeder images, the activations of each hidden node can be visualized as a 2D image. The reconstruction errors from the UFO are used to update the primary parameters (Section~\ref{sec:ufo-augmented-training}), and select reconstruction errors update the UFO parameters (Section~\ref{sec:selective-ufo-updates}). Specifically, this figure depicts \emph{intracomponent} UFO regularization of hidden node activations, and other UFO regularization techniques are introduced in Sections~\ref{sec:intracomponent-intercomponent-ufos} and~\ref{sec:UFO-attention-regularization}. 
    }
    \label{fig:method-overview-intra}
\end{figure*}

The UFO method, depicted in Figure~\ref{fig:method-overview-intra}, augments the training of a primary network with an auxiliary UFO network. The primary network is trained with gradient descent both to achieve low loss on some task objective and to make its internal representations compressible by the UFO network. The UFO itself is trained with the intent to model and amplify regular representations from the primary network. While the UFO could be implemented with many different architectures, in this paper the UFO is a non-linear autoencoder.

\subsection{Motivation}
\label{sec:motivation}

\begin{figure}[!htbp]
    \centering
    \includegraphics[width=\textwidth]{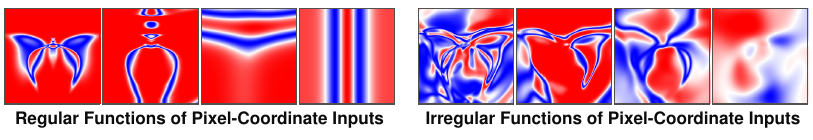}
    \caption{\textbf{UFO motivation --- regular versus irregular hidden node functions of pixel-coordinate inputs.} Each square shows the activations of an individual hidden node as a function of the network's pixel-coordinate inputs. 
    Hidden node activation maps from the original Picbreeder butterfly network (left and in Figure~\ref{fig:picbreeder-butterfly}) are far more regular than those from a network trained with conventional gradient descent to output the same butterfly (right and in Figure~\ref{fig:sgd-butterfly}). Motivated by the symmetry of individual hidden node activation maps in Picbreeder networks, UFO regularization aims to encourage each hidden node to also compute a regular function of the network inputs. 
    }
    \label{fig:regular-irregular}
\end{figure}

To motivate UFO regularization, Figure~\ref{fig:regular-irregular} shows that hidden node activation maps from the original Picbreeder butterfly network (Figure~\ref{fig:picbreeder-butterfly} and left side of Figure~\ref{fig:regular-irregular}) are far more regular than those from a network trained with conventional gradient descent to output the same image (right side of Figure~\ref{fig:regular-irregular} and Figure~\ref{fig:sgd-butterfly}).  Similarly, hidden node activation maps from a transformer trained on division modulo $97$ are more regular when the network has grokked than when it has only fit the training data (Figure~\ref{fig:trajectory-map}). With the intent for gradient descent to find such regular representations, UFO-augmented training asks that a primary network produce hidden node activations that are compressible by the UFO autoencoder. Intuitively, if the activations of each hidden node are more easily reconstructed by the UFO autoencoder, then these hidden node activations are more likely to be regular like those in Picbreeder networks.  

\subsection{Training with a UFO}
\label{sec:ufo-training}

This section explains UFO regularization in three parts: how a UFO $g_\phi$ regularizes a primary network $f_\theta$,
how a warm-up schedule prevents the regularization pressure from being too strong early in training,
and how the UFO is updated selectively in an attempt to avoid reducing regularization pressure on the primary network's irregular representations.
The UFO regularization approach described first aims to directly encourage individual hidden node activation maps like those found in Picbreeder networks. Later sections introduce additional kinds of UFO regularization.

\subsubsection{Training a Primary Network with UFO Regularization}
\label{sec:ufo-augmented-training}

The main idea of UFO regularization is to encourage a primary network $f_\theta$ to fit its training data while making its representations compressible by a UFO autoencoder $g_\phi$. At each training step $t$, $f_\theta$'s representations are extracted and reconstructed by $g_\phi$. To begin, let a representation $r_k$ from $f_\theta$ capture the activations of an individual hidden node across all examples in a training batch $\mathcal{B}$, where each hidden node's activations are $z$-score normalized independently over $\mathcal{B}$ to have zero mean and unit variance. For networks trained to output Picbreeder images, these $r_k$ capture the activations of an individual hidden node over all pixel-coordinate inputs, and thus correspond to a hidden node activation map image (Figure~\ref{fig:cppn-image-visualizations}). Associated metadata $m_k$ is also created for each $r_k$. For example, $m_k$ can be a one-hot vector describing the layer from which $r_k$ was extracted. The UFO autoencoder $g_\phi$ then reconstructs $r_k$ as $\widehat{r}_k = g_\phi([m_k;r_k])$. Let $\mathcal{L}_\mathrm{aux}(\theta, \phi; \mathcal{B})$ be the mean-squared reconstruction error of the UFO autoencoder $g_\phi$ on the representations $r_k$ extracted from $f_\theta$ during batch $\mathcal{B}$,
\begin{equation}
\label{eq:ufo-loss}
\mathcal{L}_{\mathrm{aux}}(\theta, \phi; \mathcal{B})
= \frac{1}{Kn}
\sum_{k=1}^{K}
\left\lVert \widehat{r}_k - r_k \right\rVert_2^2,
\end{equation}
where $K$ is the total number of representations $r_1, ..., r_K$ extracted from $f_\theta$ for a batch $\mathcal{B}$ and $n = |\mathcal{B}|$ is the number of examples in the batch. 

$\mathcal{L}_\mathrm{aux}$ then serves as an auxiliary loss for the primary network $f_\theta$. The primary network $f_\theta$ is trained to minimize the composite loss

\begin{equation}
\label{eq:composite-loss}
\mathcal{L}_{\mathrm{composite}}(\theta, \phi; \mathcal{B})
=
\mathcal{L}_{\mathrm{task}}(\theta;\mathcal{B})
+
\lambda(t)\,
\mathcal{L}_{\mathrm{aux}}(\theta, \phi; \mathcal{B}),
\end{equation}

where $\mathcal{L}_{\mathrm{task}}(\theta;\mathcal{B})$ is a task loss (e.g.\ mean-squared error of pixels for image regression, cross entropy for next token prediction etc.) and $\lambda(t)$ determines a UFO regularization schedule. By minimizing $\mathcal{L}_{\mathrm{composite}}$, $f_\theta$ is encouraged to achieve low task loss $\mathcal{L}_{\mathrm{task}}$ with representations that $g_\phi$ can compress (i.e.\ low $\mathcal{L}_{\mathrm{aux}}$). In contrast to a bottleneck layer within $f_\theta$, a hard constraint that would risk overly constraining learning, $\mathcal{L}_{\mathrm{aux}}$ is a soft constraint on $f_\theta$'s activations. Because $\mathcal{L}_{\mathrm{aux}}$ is just a soft constraint, the UFO autoencoder $g_\phi$ can feature a small, aggressive bottleneck, and learning for $f_\theta$ remains tractable. Furthermore, because each $r_k$ is normalized to have zero mean and unit variance, $f_\theta$ cannot trivially minimize $\mathcal{L}_{\mathrm{aux}}$ by shrinking its activation magnitude.\footnote{Related representation regularizers similarly attempt to prevent shrinking activation norms \citep{bhalodia2019coopsubnetcooperatingsubnetworkdatadriven, sanyal2020robustnessdeeplowrankrepresentations, lai2025saereg}.} Instead, $f_\theta$ must change the shape of its activations rather than just their scale. 

When $f_\theta$ is updated using $\mathcal{L}_{\mathrm{composite}}$, the UFO parameters $\phi$ are held fixed and are not updated. Although the UFO parameters $\phi$ are frozen, gradients still pass through $g_\phi$ to update the primary parameters $\theta$. The UFO parameters $\phi$ are updated separately after this primary update.

\subsubsection{UFO Regularization Warm-up}
\label{sec:ufo-warmup}

Strong UFO regularization early in training risks preventing the primary network $f_\theta$ from fitting the training data at all. To mitigate this risk, UFO regularization is ``warmed up.'' The UFO regularization schedule $\lambda(t)$ starts at zero and then is raised gradually to the steady-state level $\lambda_{\mathrm{set}}$. During this $\lambda(t)$ warm-up, the primary network $f_\theta$ can begin to fit the training data with little or no UFO regularization pressure.

\subsubsection{Selective UFO Updates}
\label{sec:selective-ufo-updates}

If the UFO $g_\phi$ were updated to better reconstruct \emph{all} activations from the primary network $f_\theta$, $g_\phi$ would risk overly accommodating $f_\theta$'s least regular activations. Because irregular activations are likely to have high reconstruction error, they could potentially dominate the UFO update and incentivize $g_\phi$ to accommodate them more than anything else.  In effect, updating the UFO parameters $\phi$ on all activations risks training away the exact regularization pressure that $g_\phi$ is intended to impose. Rather than adapting to irregular activations itself, $g_\phi$ should pressure $f_\theta$ to make its irregular activations more compressible. 

In an attempt to avoid training away the UFO's regularization pressure on irregular activations, the UFO parameters $\phi$ are updated \emph{selectively}. That is, for each hidden node, training examples that lead to activations with high reconstruction error are not used to update the UFO parameters $\phi$. Instead, $f_\theta$ must adjust so that these activations are compressible by $g_\phi$. 

To make selective UFO updates more precise, note that each entry $r_k[i]$ in a representation vector $r_k$ corresponds to an activation from a hidden node for a training example $x_i \in \mathcal{B}$. A selectivity schedule $s(t) \in [0,1]$ sets the number of examples used to update the UFO parameters $\phi$ per $r_k$ as $n_{\mathrm{sel}} = s(t) \cdot n $ (rounded to the nearest integer). For each $r_k$ independently, a binary mask $M_{k,t} \in \{0,1\}^n$ selects the $n_{\mathrm{sel}}$ entries $r_k[i]$ with the lowest reconstruction error $(r_k[i] - \widehat{r}_k[i])^2$. The UFO parameters $\phi$ are then updated to minimize the selective loss
\begin{equation}
\label{eq:selective-loss}
\mathcal{L}_{\mathrm{select}}(\theta, \phi; \mathcal{B})
= \frac{1}{K n_{\mathrm{sel}}}
\sum_{k=1}^{K}
\left\lVert M_{k,t} \odot \bigl(\widehat{r}_k - r_k\bigr) \right\rVert_2^2,
\end{equation}
where $\odot$ denotes elementwise multiplication. Note that the mask $M_{k,t}$ only determines which entries $r_k[i]$ contribute to $\mathcal{L}_{\mathrm{select}}$, and the UFO $g_\phi$ still receives the full representation $r_k$ as input. 

In most experiments below, the selectivity schedule $s(t)$ is gradually lowered over training. By starting off less selective, $g_\phi$ is intended to develop a better initial model of $f_\theta$'s representations across a wider set of inputs before becoming more restrictive.

When this selective update is applied to the UFO parameters $\phi$, a stop-gradient is placed on the representations $r_k$ so that the primary network $f_\theta$ is not updated a second time. At each step $t$ there are two updates: a primary update that trains $f_\theta$ to lower its task loss with hidden node activations that are compressible by $g_\phi$, and a selective UFO update that is intended to train $g_\phi$ to model representational regularity in $f_\theta$ while preventing $g_\phi$ from reducing regularization pressure on $f_\theta$'s irregular activations. 

\subsection{Intracomponent and Intercomponent UFOs}
\label{sec:intracomponent-intercomponent-ufos}

\begin{figure*}[!htbp]
    \centering
    \begin{subfigure}[b]{0.537\textwidth}
        \centering
        \includegraphics[width=\textwidth]{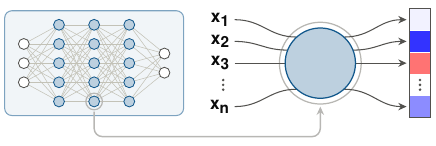}
        \caption{Intracomponent}
        \label{fig:intra-method}
    \end{subfigure}%
    \hfill
    \begin{subfigure}[b]{0.363\textwidth}
        \centering
        \includegraphics[width=\textwidth]{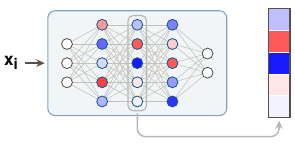}
        \caption{Intercomponent}
        \label{fig:inter-method}
    \end{subfigure}
    \caption{\textbf{Constructing intracomponent and intercomponent representations.} In panel (a), the activations of a particular hidden node are captured across inputs $x_1, x_2, \ldots, x_n$ and after normalization are transformed into an intracomponent representation $r_{k,\mathrm{intra}}$. When $f_\theta$ is trained with minibatches, the hidden node's activations may be further transformed before becoming $r_{k,\mathrm{intra}}$, as explained later in Section~\ref{sec:grokking-experimental-setup}. In panel (b), a layer's activations are captured for a particular training input $x_i$ and are normalized to become an intercomponent representation $r_{k,\mathrm{inter}}$.}
    \label{fig:intra-inter-method}
\end{figure*}

\begin{figure*}[!htbp]
    \centering
\includegraphics{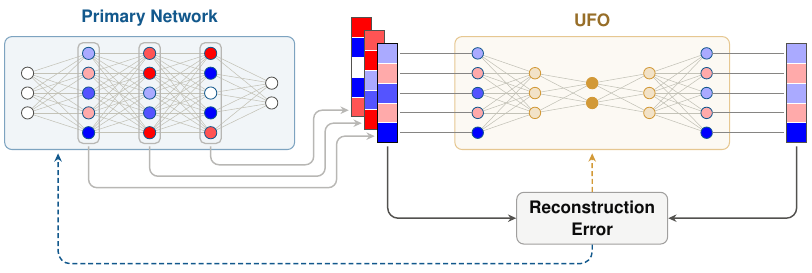}
    \caption{\textbf{Intercomponent UFO regularization}. For each training example $x_i \in \mathcal{B}$, each hidden layer's activations are extracted and normalized to form representations $r_{k,\mathrm{inter}}$. These representations $r_{k,\mathrm{inter}}$ are then input into the UFO autoencoder $g_\phi$. The reconstruction errors from the UFO $g_\phi$ are used to update the primary parameters $\theta$, and select reconstruction errors update the UFO parameters $\phi$.
    }
    \label{fig:method-overview-inter}
\end{figure*}

So far only \emph{intracomponent} UFO regularization has been described. The term ``intracomponent'' refers to the fact that a representation $r_{k, \mathrm{intra}}$ captures activations \emph{within} one hidden node for all $x_i \in \mathcal{B}$. In addition to intracomponent UFO regularization, \emph{intercomponent} UFO regularization is also possible, wherein each representation $r_{k, \mathrm{inter}}$ instead captures a layer's activations \emph{across} hidden nodes for a given $x_i \in \mathcal{B}$. Intracomponent and intercomponent representation extraction is depicted in Figure~\ref{fig:intra-inter-method}. In subsequent sections, we refer to a representation as $r_k$ in general, and as $r_{k, \mathrm{intra}}$ or $r_{k, \mathrm{inter}}$ when the distinction between intracomponent and intercomponent representations matters. With intercomponent UFO regularization, depicted in Figure~\ref{fig:method-overview-inter}, $f_\theta$ is encouraged to produce representations $r_{k,\mathrm{inter}}$ that are compressible by the UFO autoencoder $g_\phi$. 

The intracomponent approach is appealing because it focuses explicitly on the activation patterns of individual nodes, which were the clues to UFR in Picbreeder. 
However, a shortcoming of intracomponent UFO regularization is its limited scalability to large domains. Because each intracomponent representation $r_{k,\mathrm{intra}}$ is intended to model a hidden node's activations across inputs in the domain, the size of $r_{k,\mathrm{intra}}$ is tied to the number of inputs it covers. As a result, intracomponent UFO regularization (as currently conceived) is impractical for large domains. Intercomponent UFO regularization is designed to address this limitation. Each intercomponent representation $r_{k, \mathrm{inter}}$ captures a layer's activations for just one input in the domain, so although the number of representations $r_{k, \mathrm{inter}}$ grows with the size of the domain, each $r_{k, \mathrm{inter}}$ is always only the size of a layer's width. Because the size of each intercomponent representation $r_{k, \mathrm{inter}}$ is not tied to the size of the domain, intercomponent UFO regularization is more scalable to large domains. 

To better understand the relationship between intracomponent and intercomponent representations, let $R_\theta^{(\ell)}(\mathcal B)\in\mathbb R^{n\times c}$ capture the activations of $c$-many hidden nodes in layer $\ell$ of $f_\theta$\footnote{In all experiments in this paper, every layer of $f_\theta$ has the same width $c$, so each $r_{k,\mathrm{inter}}$ has length $c$ and a single UFO $g_\phi$ can reconstruct intercomponent representations from every layer.} over all $n$ training examples in $\mathcal{B}$, where each column of $R_\theta^{(\ell)}(\mathcal B)$ is $z$-score normalized to have zero mean and unit variance. A \emph{column} of $R_\theta^{(\ell)}(\mathcal B)$ corresponds to an intracomponent representation vector $r_{k, \mathrm{intra}}$, and a \emph{row} of $R_\theta^{(\ell)}(\mathcal B)$ corresponds to an intercomponent representation vector $r_{k, \mathrm{inter}}$. In this way, intracomponent and intercomponent regularization are two different ways of regularizing the same hidden node activation matrix $R_\theta^{(\ell)}(\mathcal B)$. 

As in the intracomponent case, intercomponent UFO updates are applied selectively so that training examples $x_i$ that lead to high reconstruction error are not used to update the UFO parameters $\phi$. In the intracomponent case, each training example $x_i$ maps to a single entry $r_{k,\mathrm{intra}}[i]$, but in the intercomponent case, each training example $x_i$ maps to an entire representation $r_{k,\mathrm{inter}}$. So for intercomponent UFO regularization, within each layer only the $n_{\mathrm{sel}}$ representations $r_{k,\mathrm{inter}}$ with the lowest reconstruction error $\lVert \widehat{r}_k - r_k \rVert_2^2$ are used to update $\phi$. That means that $n_{\mathrm{sel}}$ training examples are used per layer, just as $n_{\mathrm{sel}}$ are used per hidden node in the intracomponent case.

Two small changes adapt $\mathcal{L}_\mathrm{aux}$ (Equation~\ref{eq:ufo-loss}) and $\mathcal{L}_\mathrm{select}$ (Equation~\ref{eq:selective-loss}) to intercomponent regularization. First, each mask $M_{k,t} \in \{0,1\}^c$ now selects an entire representation rather than individual entries, so a mask is all ones if $r_{k,\mathrm{inter}}$ is selected and all zeros otherwise. Second, the normalizing constants are adjusted so that both $\mathcal{L}_\mathrm{aux}$ (Equation~\ref{eq:ufo-loss}) and $\mathcal{L}_\mathrm{select}$ (Equation~\ref{eq:selective-loss}) remain mean-squared errors. See Appendix~\ref{app:normalizers} for more details on normalizing constants.

\subsection{Regularizing Attention with UFOs}
\label{sec:UFO-attention-regularization}
In addition to regularizing hidden node activations, intracomponent and intercomponent UFOs can regularize attention weights. In fact, the term \emph{component} was chosen because it can refer to both hidden nodes and attention edges, where an \emph{attention edge} describes the query-key connection between two fixed sequence positions in a given attention head. For example, an attention edge corresponding to the connection between the \texttt{=} token and \texttt{x} token is shown in the third row of Figure~\ref{fig:trajectory-map}, and regularity can be seen incrementally taking shape for this attention edge, similarly to the hidden nodes depicted in the same figure. With intracomponent UFO regularization of attention, a representation $r_{k,\mathrm{intra}}$ captures the attention weights for a given attention edge across all $x_i \in \mathcal{B}$; in the intercomponent attention case, a representation $r_{k, \mathrm{inter}}$ captures the attention weights across multiple attention edges in a given layer of $f_\theta$ for a training example $x_i \in \mathcal{B}$.

\subsection{Summary}
\label{sec:method-summary}

UFOs are motivated by the regular representations of Picbreeder networks and grokked transformers. With a UFO $g_\phi$, a primary network $f_\theta$ is trained to minimize a task loss while making its representations compressible by $g_\phi$. Because these representations are normalized before being passed to $g_\phi$, $f_\theta$ must make its representations compressible by adjusting their shape rather than merely their scale. The UFO regularization schedule $\lambda(t)$ is warmed up with the intent that early regularization pressure does not disrupt task learning. The UFO $g_\phi$ is updated selectively with the intent that it does not accommodate $f_\theta$'s  irregular representations, instead pressuring $f_\theta$ to adjust these irregular representations. UFOs can regularize \emph{intracomponent} representations $r_{k,\mathrm{intra}}$, which capture a single hidden node's activations across all training examples in a batch $\mathcal{B}$, or \emph{intercomponent} representations $r_{k,\mathrm{inter}}$, which capture a layer's activations for a single training example. In addition to hidden node activations, UFOs can also regularize attention weights. A general algorithm that applies to both intracomponent and intercomponent UFOs that regularize either hidden node activations or attention weights is given in Algorithm~\ref{alg:ufo}.

\begin{algorithm}[H]
\caption{Training a primary network $f_\theta$ with a UFO $g_\phi$. \label{alg:ufo}}
\begin{algorithmic}[1]
\setlength{\itemsep}{2pt}
\Require Primary network $f_\theta$, UFO $g_\phi$, training data $\mathcal{D}$, regularization schedule $\lambda(t)$, selectivity schedule $s(t)$, number of steps $T$
\For{$t = 1, \ldots, T$}
    \State Sample a training batch $\mathcal{B} \subseteq \mathcal{D}$ with $n = |\mathcal{B}|$ examples
    \State Extract normalized representations $r_1, \ldots, r_K$ from $f_\theta$ on $\mathcal{B}$, with metadata $m_k$ \Comment{Sec.~\ref{sec:ufo-augmented-training}}
    \State Reconstruct $\widehat{r}_k \gets g_\phi([m_k; r_k])$ for all $k$
    \Statex \hspace{\algorithmicindent}\textbf{Primary update}
    \State Compute task loss $\mathcal{L}_{\mathrm{task}}$ and auxiliary loss $\mathcal{L}_{\mathrm{aux}}$ \Comment{Eq.~\eqref{eq:ufo-loss}}
    \State $\mathcal{L}_{\mathrm{composite}} \gets \mathcal{L}_{\mathrm{task}} + \lambda(t)\,\mathcal{L}_{\mathrm{aux}}$ \Comment{Eq.~\eqref{eq:composite-loss}}
    \State Update $\theta$ using $\nabla_\theta \mathcal{L}_{\mathrm{composite}}$, with $\phi$ held fixed
    \Statex \hspace{\algorithmicindent}\textbf{Selective UFO update}
    \State $n_{\mathrm{sel}} \gets  s(t) \cdot n$
    \State Compute masks $M_{k,t}$ selecting the $n_{\mathrm{sel}}$ lowest-error reconstructions \Comment{Sec.~\ref{sec:selective-ufo-updates}}
    \State Compute selective loss $\mathcal{L}_{\mathrm{select}}$ with a stop-gradient on each $r_k$ \Comment{Eq.~\eqref{eq:selective-loss}}
    \State Update $\phi$ using $\nabla_\phi \mathcal{L}_{\mathrm{select}}$, with $\theta$ held fixed
\EndFor
\end{algorithmic}
\end{algorithm}

\section{Picbreeder Experiments}
\label{sec:picbreeder-experiments}

The idea behind the experimental regime in this paper is first to establish the qualitative, visually recognizable improvements that the UFO produces.  Picbreeder images are a useful domain for such a demonstration because there are already not only examples of networks with UFR for Picbreeder images from the original Picbreeder itself, but also evidence that conventional gradient descent on its own does not produce such UFR \citep{kumar2025questioningrepresentationaloptimismdeep}.

Once the qualitative benefits are established, the experiments in Section~\ref{sec:grokking-experiments} turn to a quantitative analysis in grokking that establishes that the more regular representation seen in the Picbreeder domain indeed translates to downstream impact in a domain where representation is likely to matter.

\subsection{Picbreeder Experimental Setup}
\label{sec:picbreeder-experimental-setup}

To demonstrate the qualitative representational improvements induced by UFOs, CPPNs are trained to render the Picbreeder butterfly image without regularization, with intracomponent UFO regularization, and with intercomponent UFO regularization. All of these CPPNs are trained for 50,000 steps with the Adam optimizer \citep{kingma2017adammethodstochasticoptimization}. Loss relative to the target butterfly image is measured as pixel-level mean-squared error, and the networks are trained with full batches so at each step all $256 \times 256$ pixels in the butterfly image contribute to the loss. To facilitate comparison to the original Picbreeder network, networks use a layerized MLP architecture that is computationally equivalent to the Picbreeder-evolved architecture for the butterfly. Layerization for Picbreeder networks is described in \citet{kumar2025questioningrepresentationaloptimismdeep}; a similar process was first described in chapter 3.2 of \citet{budenbender2021revisiting}. This layerized MLP network is then the primary network $f_\theta$. 

In both the intracomponent and intercomponent cases, the UFO network $g_\phi$ is a non-linear autoencoder. In the intracomponent case, $g_\phi$ has a $2$-dimensional bottleneck layer, and a $128$-dimensional hidden layer for both the encoder and decoder. In the intercomponent case, $g_\phi$ has a $4$-dimensional bottleneck layer and a $64$-dimensional hidden layer for both the encoder and decoder. In both cases, the UFO regularization schedule $\lambda(t)$ is ``warmed up'' so that $\lambda(t) = 0$ for the first $1$,$000$ steps, $\lambda(t)$ is raised linearly to $\lambda_{\mathrm{set}}$ from step $1$,$000$ to $5$,$000$, and $\lambda(t) = \lambda_{\mathrm{set}}$ thereafter. In the intracomponent case, $\lambda_{\mathrm{set}} = 0.1$, and in the intercomponent case $\lambda_{\mathrm{set}} = 0.3$. Both UFO-regularized networks use a selectivity schedule $s(t)$. In the intracomponent case, $s(t) = 0.25$ is constant throughout training. In the intercomponent case, $s(t)$ is lowered linearly over the first $20$,$000$ steps from $1$ to $0.1$ and $s(t) = 0.1$ thereafter. All hyperparameters were determined through empirical trial and error testing. 

In the intracomponent case, a representation vector $r_{k, \mathrm{intra}}$
is extracted for each hidden node and captures that hidden node's activations for all $256 \times 256$ pixel coordinates in the domain. As a result, each $r_{k,\mathrm{intra}}$ corresponds to a hidden node activation map, as depicted in Figure~\ref{fig:cppn-image-visualizations}. In the intercomponent case, for each layer in $f_\theta$ and for each pixel coordinate in the $256 \times 256$ input domain, a representation vector $r_{k, \mathrm{inter}}$ is extracted capturing that layer's hidden node activations at that pixel coordinate. 

In both cases, metadata $m_k$ is created for each $r_k$. In the intracomponent case, $m_{k, \mathrm{intra}}$ is a vector indexing both the layer from which $r_{k, \mathrm{intra}}$ is extracted and the activation function for the hidden node associated with $r_{k, \mathrm{intra}}$. The layerized MLP architecture that is trained to output the butterfly uses the heterogeneous activation functions from the original Picbreeder-evolved architecture, including sigmoid, sine, Gaussian, and the identity. To assist the intracomponent UFO $g_\phi$ in compressing $r_{k,\mathrm{intra}}$, this metadata $m_{k, \mathrm{intra}}$ is inserted at both the encoder and decoder. In the intercomponent case, $m_{k, \mathrm{inter}}$ is a one-hot vector giving an index for the layer from which $r_{k, \mathrm{inter}}$ was extracted, and $m_{k, \mathrm{inter}}$ is only passed to the encoder. 

\subsection{Picbreeder Results}
\label{sec:picbreeder-results}

\begin{figure*}[!htbp]
    \centering
    \includegraphics[
        width=\textwidth,
        height=\featuremapheight,
        keepaspectratio
    ]{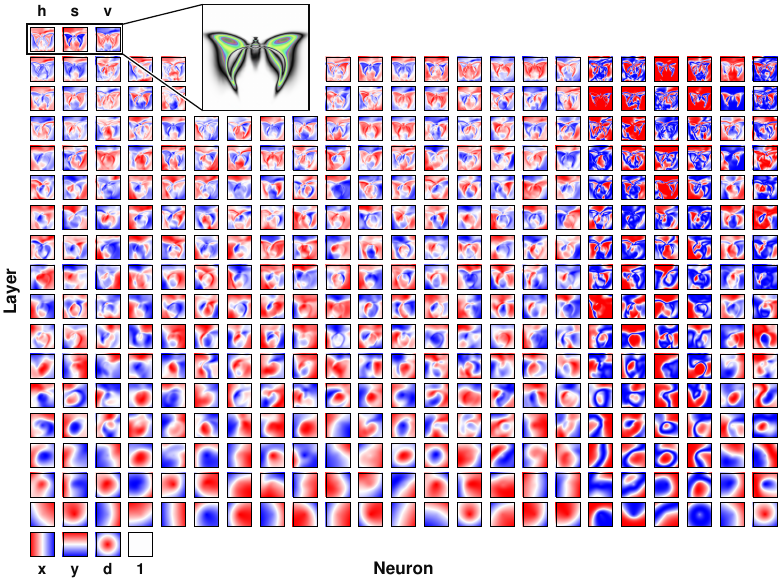}
    \caption{\textbf{Representations of a conventional gradient descent baseline network trained to output the butterfly.}  Hidden node activations are irregular and high-frequency and fail to capture the symmetries of the butterfly. This visualization follows the same layerized architecture and visualization scheme as the original Picbreeder butterfly network shown in Figure~\ref{fig:picbreeder-butterfly}.}
    \label{fig:sgd-butterfly}
\end{figure*}

\begin{figure}[p]
    \centering
    \vspace*{-1cm}
    \begin{subfigure}{\textwidth}
        \centering
        \includegraphics[
            width=\textwidth,
            height=\featuremapheight,
            keepaspectratio
        ]{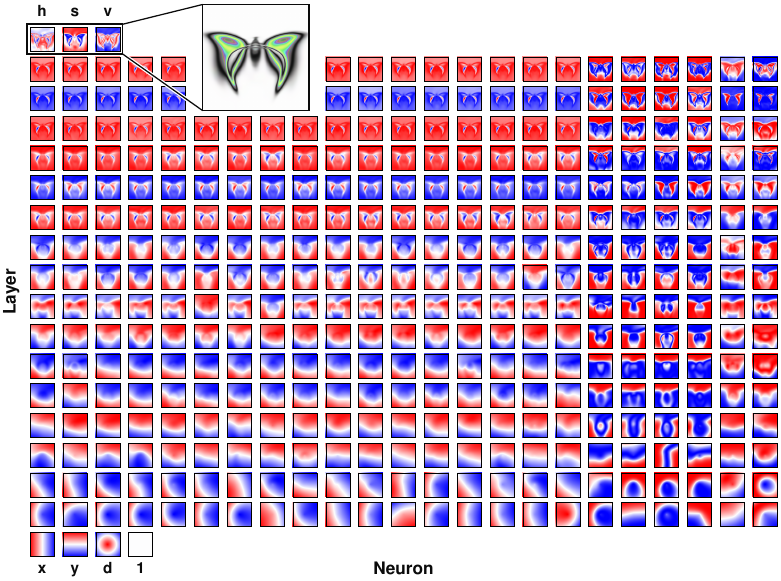}
        \caption{Intracomponent}
        \label{fig:intracomponent-ufo-butterfly}
    \end{subfigure}

    \begin{subfigure}{\textwidth}
        \centering
        \includegraphics[
            width=\textwidth,
            height=\featuremapheight,
            keepaspectratio
        ]{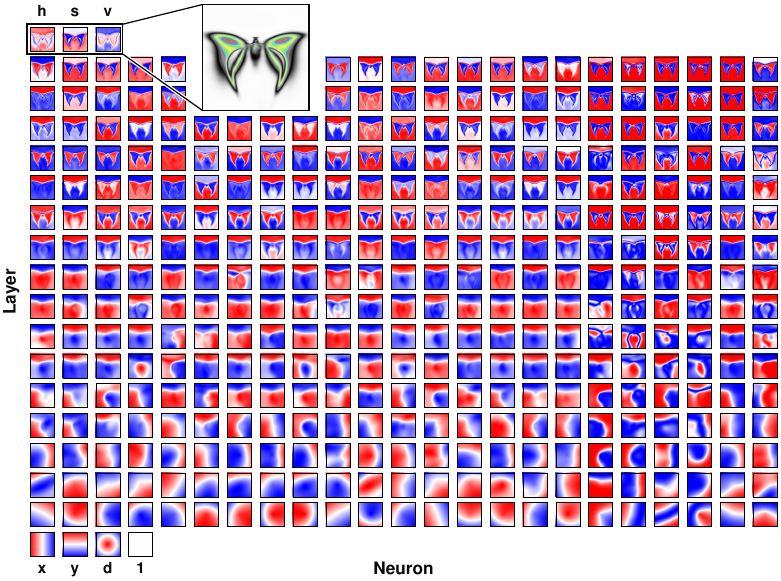}
        \caption{Intercomponent}
        \label{fig:intercomponent-ufo-butterfly}
    \end{subfigure}

    \vspace{0.8em}
    \caption{\textbf{Representations of UFO-regularized networks trained to output the butterfly.} Notice that the intracomponent UFO-regularized network (a) features many redundant, regular hidden node activation maps, and the intercomponent UFO-regularized network (b) has many hidden nodes that emphasize the butterfly's main shape. These networks follow the same layerized architecture and visualization scheme as the original Picbreeder butterfly network shown in Figure~\ref{fig:picbreeder-butterfly}.}
    \label{fig:ufo-butterfly-feature-maps}
\end{figure}

Figure~\ref{fig:sgd-butterfly} shows the representations of a representative conventional gradient descent baseline network. In keeping with the results from \citet{kumar2025questioningrepresentationaloptimismdeep}, the conventional gradient descent baseline exhibits FER --- hidden node activations are irregular and high-frequency, and the network's representations fail to consistently capture the salient symmetries of the butterfly. Although UFO-regularized results are not always consistent (the random seed used for network initialization can influence representational quality and poorly tuned hyperparameters can lead to underfitting), representations shown in Figures~\ref{fig:intracomponent-ufo-butterfly} and~\ref{fig:intercomponent-ufo-butterfly} illustrate successful intracomponent and intercomponent UFO-regularized networks, respectively. These representations are particularly interesting because \citet{kumar2025questioningrepresentationaloptimismdeep} call into question the very possibility of acquiring good representations of Picbreeder images with gradient descent. In contrast to the conventional gradient descent baseline, the intracomponent and intercomponent UFO-regularized networks' hidden node activations better respect the regularities of the butterfly --- select nodes construct prominent, regular features of the butterfly and are more likely to have symmetric activation maps than the conventional gradient descent baseline. The UFO-regularized networks also have distinct styles of their own: the intracomponent network has many redundant and regular hidden nodes while the intercomponent network's hidden nodes emphasize the butterfly's main shape. 

In both cases, and in the intracomponent case in particular, $g_\phi$ is asked to radically compress $f_\theta$'s activations. Importantly, this demand for radical compression is only a soft constraint on $f_\theta$, and $f_\theta$ is still able to output the target butterfly despite $g_\phi$'s aggressive bottleneck. However, both UFO-regularized networks slightly underfit the small butterfly body compared to the unregularized network, suggesting that during the optimization process the network traded off lower pixel mean-squared error for a lower UFO regularization loss. For its part, the UFO $g_\phi$ reconstructs the activations of $f_\theta$ reasonably well. To illustrate the fidelity of the UFO's reconstruction, Figure~\ref{fig:reconstructed-intracomponent} and Figure~\ref{fig:reconstructed-intercomponent} in Appendix~\ref{app:picbreeder} visualize the UFO's reconstructed activations for the intracomponent and intercomponent regularized CPPNs shown in Figures~\ref{fig:intracomponent-ufo-butterfly} and~\ref{fig:intercomponent-ufo-butterfly}. Appendix~\ref{app:picbreeder} also shows additional representations from UFO-regularized CPPNs that output the butterfly.  

Overall, the sharp qualitative difference between representations in UFO-regularized networks and conventional gradient descent networks hints at the potential for UFOs to enable us to investigate the implications of UFR for different kinds of learning. In that spirit, the next section explores the UFO's impact on grokking.

\section{Grokking Experiments}
\label{sec:grokking-experiments}

 To complement the qualitative results presented in Section~\ref{sec:picbreeder-experiments}, this section shows that UFOs can significantly accelerate grokking of modular division problems, providing concrete, quantitative evidence for the potential of UFOs in inducing good representations. More details on the grokking experiments are in Appendix~\ref{app:grokking}. 

\subsection{Grokking Experimental Setup}
\label{sec:grokking-experimental-setup}

To further test the UFO's influence on representation learning, a 2-layer, decoder-only transformer $f_\theta$ is trained to grok division modulo 97. This task, modulus, and architecture mirror a canonical setting in which grokking was first reported by \citet{power2022grokkinggeneralizationoverfittingsmall}. Division modulo 97 defines a fixed set of $97 \cdot 96 = 9$,$312$ problems of the form $x / y = z\pmod{97}$, where $x, z \in \mathbb{Z}_{97}$ and $y \in \mathbb{Z}_{97}\setminus \{0\}$. After defining tokens for each element of $\mathbb{Z}_{97}$, as well as for \texttt{/}, \texttt{=}, and a special end-of-sequence token \texttt{<|eos|>}, $f_\theta$ is trained to predict the correct next token \texttt{z} when given the token sequence \texttt{<|eos|> x / y =} by minimizing cross-entropy loss. In the following experiments, $f_\theta$ is trained on half of the total $9$,$312$ modular division problems and its accuracy on the held-out, validation half of problems is measured every $1$,$000$ training steps. A key metric is the ``grok step,'' the first recorded step at which $f_\theta$ achieves $\geq 99 \%$ validation accuracy. 

The primary network $f_\theta$ is trained with minibatches of $512$ problems per step and AdamW is the optimizer \citep{loshchilov2019decoupledweightdecayregularization} (see Appendix~\ref{app:grokking} for architecture and optimizer details). Because grokking of modular arithmetic problems can be accelerated significantly with weight decay \citep{power2022grokkinggeneralizationoverfittingsmall}, $f_\theta$ is trained with UFO-based regularization both with and without weight decay. All experiments with weight decay use a weight decay of $0.03$. This weight decay value is chosen because it is on the higher end of what might be used in practice, and when paired with a training data fraction of $50\%$ it causes $f_\theta$ to grok reliably in approximately $100$,$000$ training steps without additional UFO regularization. With this consistent grokking timing with weight decay alone, the speedup from adding UFO regularization to weight decay can be measured cleanly.

Intracomponent and intercomponent UFOs are tested with two representation modalities, MLP activations and attention weights, for a total of four UFO types: intracomponent MLP, intracomponent attention, intercomponent MLP, and intercomponent attention. For all UFO types, the UFO $g_\phi$ is a non-linear autoencoder with one hidden layer for the encoder and decoder and a $2$-dimensional bottleneck, and $g_\phi$ is trained with the Adam optimizer and no weight decay. In all cases, $f_\theta$'s representations $r_k$ are extracted and input into $g_\phi$ when $f_\theta$ is processing the \texttt{=} token to predict the next token \texttt{z} in an equation \texttt{x / y = z}.
For both attention cases, only attention edges from a \texttt{=}-query row are regularized (see Appendix~\ref{app:grokking} for more details). The UFO regularization hyperparameters and autoencoder architectures for each treatment were determined through empirical trial and error testing. 

To set up intracomponent UFO regularization, every division modulo $97$ problem is assigned a fixed position between $0$ and $9$,$311$. A hidden node's activation or an attention edge's weight is assigned to its corresponding position for each problem in the batch $\mathcal{B}$, and all positions corresponding to a problem not in $\mathcal{B}$ are filled with zeros; this process yields $r_{k, \mathrm{intra-MLP}}$ or $r_{k, \mathrm{intra-attn}}$, respectively. Reconstruction losses are computed using only the coordinates of $r_{k, \mathrm{intra}}$ that correspond to component responses and not the $0$-filled positions. In the intracomponent MLP and attention cases, $g_\phi$ is an autoencoder with a $128$-dimensional hidden layer for both the encoder and decoder. In both intracomponent cases, the selectivity schedule $s(t)$ decreases linearly from $0.50$ to $0.10$ over the first $33$,$333$ steps, and then $s(t) = 0.10$ thereafter.

In the intercomponent MLP case, each representation vector $r_{k, \mathrm{inter-MLP}}$ captures a layer's normalized MLP activations for one problem in the training batch $\mathcal{B}$. In the intercomponent attention case, a representation vector $r_{k, \mathrm{inter-attn}}$ concatenates the attention weights at the \texttt{=}-query row across all heads in a layer, also for one problem in $\mathcal{B}$. The representation vector $r_{k, \mathrm{inter-attn}}$ is significantly smaller than $r_{k, \mathrm{inter-MLP}}$, so the intercomponent attention UFO uses a smaller, $8$-dimensional hidden layer for both the encoder and the decoder while the intercomponent MLP UFO uses a $128$-dimensional hidden layer. For both intercomponent cases, the selectivity schedule $s(t)$ decreases linearly from $0.25$ to $0.10$ over the first $2$,$000$ steps, and then $s(t) = 0.10$ thereafter.

For all UFO treatments, the regularization schedule $\lambda(t)$ is $0$ from step $0$--$250$, increases linearly over steps $250$--$500$ to some set amount $\lambda_{\mathrm{set}}$, and then $\lambda(t) = \lambda_{\mathrm{set}}$ thereafter. The $\lambda_{\mathrm{set}}$ values for each treatment are in Tables~\ref{tab:grokking_treatments} and~\ref{tab:grokking_unregularized}. Additionally, for all UFO treatments the metadata $m_k$ is a $2$-dimensional one-hot vector that indexes the layer from which a representation vector $r_{k}$ was extracted. Unlike in the Picbreeder intracomponent case described in Section~\ref{sec:picbreeder-experimental-setup}, this metadata $m_k$ does not also communicate the activation function type because all hidden nodes in $f_\theta$ use ReLU activations. Also, $m_k$ is only included at the encoder and is not reinserted at the bottleneck layer for the decoder. 

Three controls provide baselines for the four UFO treatments described above: a weight-decay-only baseline with no UFO regularization, a UFO network that is randomly initialized and does not update during training, and a non-selective UFO network that updates on every primary network representation at each step (i.e.\ $s(t) = 1$ for all steps $t$). The fixed and non-selective controls use the same autoencoder architecture and regularization schedule $\lambda(t)$ as the selective UFO treatments described above. All of these treatments use a weight decay of $0.03$. Each treatment is tested using eight seeds that vary the  primary parameters $\theta$, the UFO parameters $\phi$, and the minibatch data shuffling. 

Three key statistics are recorded:
(1) each treatment's median grok step over these eight seeds, (2) the grok rate, i.e.\ the fraction of seeds for which the network grokked in an allocated $150$,$000$ transformer training step budget, and (3) each treatment's median grokking speedup, computed as the weight-decay-only control's median grok step divided by the treatment's median grok step. Specifically, each median grok step is calculated using the Kaplan--Meier method (i.e.\ as the earliest step by which at least half of a treatment's eight runs grokked, which means runs that did not grok within the step budget are included in the median calculation).

To determine if UFO regularization alone can accelerate grokking, the four selective UFO treatments are also tested without weight decay. Other than removing weight decay, all other hyperparameters are the same as in the weight decay case outlined above, with the exception that instead of running for a maximum of $150$,$000$ steps, these experiments run for a maximum of $4$,$000$,$000$ steps because grokking can take longer without weight decay.  These selective UFO treatments without weight decay are compared to an unregularized control that does not use weight decay or UFO regularization, and each treatment's median grokking speedup is computed as the unregularized control's median grok step divided by the treatment's median grok step.

\subsection{Grokking Results}
\label{sec: grokking results}

As shown in Table~\ref{tab:grokking_treatments}, selective UFO treatments achieve a speedup relative to weight decay alone of $\sim 2$--$12\times$, beating all control treatments in median grok step across matched settings. Overall, that selective UFOs lead to dramatic speedups in all cases suggests both that inducing regular representation is helpful and that a selectivity schedule helps to make the process more effective. The transformer's representations also become more compressible by the UFO over training. To show how this compressibility develops alongside grokking, Appendix~\ref{app:ufo-aux-trace} plots the UFO reconstruction error $\mathcal{L}_{\mathrm{aux}}$ and the transformer's validation accuracy over training for each selective UFO treatment with weight decay, using the seed with the median grok step.

Across these treatments, regularizing attention can lead to greater grokking acceleration than regularizing MLP activations (though MLP setups still perform multiple times better than weight decay alone). This difference between attention and MLP effectiveness may suggest that attention is especially important for the general algorithm learned by $f_\theta$. Another possibility is that the smaller number of representation vectors for attention in the intracomponent case or the relatively small representation vectors in the intercomponent case make it easier for the UFO to encourage good representations.

\begin{table*}
\centering
\caption{
\textbf{Grokking results with weight decay set to 0.03 for all treatments.}
The last column reports each treatment's median speedup, computed as the 98k-step weight-decay-only baseline's median grok step divided by the treatment's median grok step. In the fixed intracomponent  attention case (marked with an asterisk) only one of the 8 seeds grokked within the budget, so the Kaplan--Meier median is undefined. The grok step for that seed is reported but the speedup is not included. 
}
\label{tab:grokking_treatments}
\renewcommand{\arraystretch}{1.25}
\setlength{\tabcolsep}{5pt}

\begin{tabularx}{\textwidth}{
    @{}
    >{\raggedright\arraybackslash}p{2.15cm}
    >{\raggedright\arraybackslash}p{1.60cm}
    >{\centering\arraybackslash}p{1.35cm}
    >{\centering\arraybackslash}X
    >{\centering\arraybackslash}p{1.65cm}
    >{\centering\arraybackslash}p{1.15cm}
    >{\centering\arraybackslash}p{1.55cm}
    @{}
}
\toprule
\textbf{Treatment} &
\textbf{Modality} &
\(\boldsymbol{\lambda}_{\mathrm{set}}\) &
\makecell[c]{\textbf{Selectivity}\\\textbf{schedule}} &
\makecell[c]{\textbf{Median}\\\textbf{grok step}} &
\makecell[c]{\textbf{Grok}\\\textbf{rate}} &
\raisebox{0.6ex}{%
    \makecell[c]{\textbf{Speedup}\\\textbf{vs.\ WD only}}
} \\
\midrule
\multicolumn{7}{@{}l}{\textbf{(a) Intracomponent}} \\
\addlinespace[2pt]

\textbf{Selective}
& MLP
& 1
& \(50\% \rightarrow 10\%\) over 33,333
& \textbf{48k}
& \textbf{7/8}
& \(\boldsymbol{2.04\times}\) \\

\textbf{Selective}
& Attention
& 3
& \(50\% \rightarrow 10\%\) over 33,333
& \textbf{8k}
& \textbf{8/8}
& \(\boldsymbol{12.25\times}\) \\

Fixed
& MLP
& 1
& --
& 92k
& 8/8
& \(1.07\times\) \\

Fixed
& Attention
& 3
& --
& 131k\(^{*}\)
& 1/8
& -- \\

Non-Selective
& MLP
& 1
& Always \(100\%\)
& 79k
& 8/8
& \(1.24\times\) \\

Non-Selective
& Attention
& 3
& Always \(100\%\)
& 38k
& 8/8
& \(2.58\times\) \\

\midrule
\multicolumn{7}{@{}l}{\textbf{(b) Intercomponent}} \\
\addlinespace[2pt]

\textbf{Selective}
& MLP
& 1
& \(25\% \rightarrow 10\%\) over 2,000
& \textbf{31k}
& \textbf{8/8}
& \(\boldsymbol{3.16\times}\) \\

\textbf{Selective}
& Attention
& 5
& \(25\% \rightarrow 10\%\) over 2,000
& \textbf{13k}
& \textbf{8/8}
& \(\boldsymbol{7.54\times}\) \\

Fixed
& MLP
& 1
& --
& 41k
& 8/8
& \(2.39\times\) \\

Fixed
& Attention
& 5
& --
& 33k
& 8/8
& \(2.97\times\) \\

Non-Selective
& MLP
& 1
& Always \(100\%\)
& 59k
& 8/8
& \(1.66\times\) \\

Non-Selective
& Attention
& 5
& Always \(100\%\)
& 41k
& 8/8
& \(2.39\times\) \\

\midrule

WD only
& --
& --
& --
& 98k
& 8/8
& \(1.00\times\) \\

\bottomrule
\end{tabularx}
\end{table*}

Notably, the selective treatments always outperform their non-selective counterparts in median grok step. As discussed in Section~\ref{sec:selective-ufo-updates}, the explanation for this disparity may be the non-selective UFO $g_\phi$ ``chasing'' the representations of the primary network $f_\theta$. That is, $g_\phi$ updates so as to achieve low error on $f_\theta$'s least compressible representations, effectively reducing the regularization pressure on these least compressible representations, rather than forcing $f_\theta$ to make these potentially troublesome representations more compressible. In the selective treatment, $g_\phi$ is instead trained with the intent to model only the most regular representations from $f_\theta$, and $f_\theta$ is thus encouraged to correct irregular representations by aligning them with the model learned by $g_\phi$. 

Perhaps surprisingly, the fixed treatment (wherein weights in $g_\phi$ do not update from their small initialization) achieves a lower median grok step than both the weight-decay-only control and the non-selective control in the intercomponent cases. One hypothesis is that even a randomly initialized UFO imposes a low-dimensional constraint that is useful in this setting. Because every representation must pass through the same fixed $2$-dimensional bottleneck, a fixed $g_\phi$ may still push $f_\theta$ to make its layer activations differ across problems in only a few shared ways. This constraint perhaps encourages a common algorithm to be applied to all problems rather than an idiosyncratic representation for each problem. However, the fixed intracomponent treatment does not lead to a notable grokking speedup for MLP activations and fails to grok within the step budget for seven of eight seeds for attention weights, suggesting that a fixed autoencoder risks imposing a constraint that hinders $f_\theta$'s learning.

\begin{figure*}[!htbp]
    \centering
\includegraphics{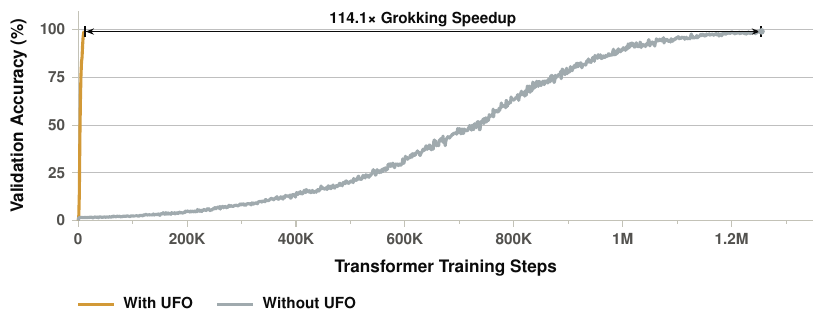}
    \caption{\textbf{Median grokking speedup caused by the fastest UFO treatment (selective intracomponent attention) when compared to an unregularized transformer.} Two transformers with matching hyperparameters are trained on division modulo $97$, one with UFO regularization (gold) and one without UFO regularization or weight decay (gray). The transformer trained with UFO regularization groks in $\sim114.1\times$ fewer transformer training steps than the unregularized transformer. This pair of runs is highlighted because each run grokked at its treatment's median grok step across eight seeds, showing the median grokking speedup induced by the fastest UFO treatment, selective intracomponent attention UFOs.} 
    \label{fig:median-speedup-combined-no-weight-decay}
\end{figure*}

Interestingly, although weight decay is a fundamental mechanism in many systems designed to grok, Table~\ref{tab:grokking_unregularized} shows that UFOs can also accelerate grokking without weight decay, leading to a one to two order-of-magnitude grokking speedup compared to the unregularized baseline. 
Figure~\ref{fig:median-speedup-combined-no-weight-decay} shows the median speedup caused by selective intracomponent attention UFOs, the fastest treatment in this setting, when used without weight decay and compared to an unregularized baseline. However, in the intracomponent case, the UFO can cause the transformer not to grok. In particular, the intracomponent MLP case only leads to grokking for one of eight seeds, suggesting that this treatment requires further adaptation to the no-weight-decay setting to recover its effectiveness. 

\begin{table*}[t]
\centering
\caption{
\textbf{Grokking results with UFO regularization alone (i.e.\ no weight decay).}
The last column reports each treatment's median speedup, computed as the 1.255M-step unregularized baseline's median grok step divided by the treatment's median grok step. In the selective intracomponent  MLP case (marked with an asterisk)
only one of the 8 seeds grokked within the budget, so the Kaplan--Meier median is undefined. The grok step for that seed is reported but the speedup is not included. 
}
\label{tab:grokking_unregularized}

\renewcommand{\arraystretch}{1.25}
\setlength{\tabcolsep}{5pt}

\begin{tabularx}{\textwidth}{
    @{}
    >{\raggedright\arraybackslash}p{2.15cm}
    >{\raggedright\arraybackslash}p{1.60cm}
    >{\centering\arraybackslash}p{1.35cm}
    >{\centering\arraybackslash}X
    >{\centering\arraybackslash}p{1.65cm}
    >{\centering\arraybackslash}p{1.15cm}
    >{\centering\arraybackslash}p{1.75cm}
    @{}
}
\toprule

\textbf{Treatment} &
\textbf{Modality} &
\(\boldsymbol{\lambda}_{\mathrm{set}}\) &
\makecell[c]{\textbf{Selectivity}\\\textbf{schedule}} &
\makecell[c]{\textbf{Median}\\\textbf{grok step}} &
\makecell[c]{\textbf{Grok}\\\textbf{rate}} &
\raisebox{0.6ex}{%
    \makecell[c]{\textbf{Speedup vs.}\\\textbf{unregularized}}
} \\

\midrule
\multicolumn{7}{@{}l}{\textbf{(a) Intracomponent}} \\
\addlinespace[2pt]

\textbf{Selective}
& MLP
& 1
& \(50\% \rightarrow 10\%\) over 33,333
& \textbf{68k}\(^{*}\)
& \textbf{1/8}
& -- \\

\textbf{Selective}
& Attention
& 3
& \(50\% \rightarrow 10\%\) over 33,333
& \textbf{11k}
& \textbf{6/8}
& \(\boldsymbol{114.09\times}\) \\

\midrule
\multicolumn{7}{@{}l}{\textbf{(b) Intercomponent}} \\
\addlinespace[2pt]

\textbf{Selective}
& MLP
& 1
& \(25\% \rightarrow 10\%\) over 2,000
& \textbf{66k}
& \textbf{8/8}
& \(\boldsymbol{19.02\times}\) \\

\textbf{Selective}
& Attention
& 5
& \(25\% \rightarrow 10\%\) over 2,000
& \textbf{16k}
& \textbf{8/8}
& \(\boldsymbol{78.44\times}\) \\

\midrule

Unregularized
& --
& --
& --
& 1.255M
& 8/8
& \(1.00\times\) \\

\bottomrule
\end{tabularx}
\end{table*}

To better contextualize the grokking speedups induced by UFOs, we compare to the grokking speedups reported by \citet{lee2024grokfastacceleratedgrokkingamplifying} and \citet{kazanskii2026tamegrokkingrepresentationgeometry} when training transformers to grok modular arithmetic problems. However, 
speedup multiples cannot be compared directly because different task and hyperparameter settings are used in each work. \citet{lee2024grokfastacceleratedgrokkingamplifying}, who accelerate grokking by amplifying the slow-varying component of gradients, report that their method reduces the number of transformer training iterations needed to achieve grokking by $\sim13.6\times$ when used without weight decay and compared to an unregularized baseline, or by $\sim20.8\times$ when applied only after the model has overfit. As seen in Table~\ref{tab:grokking_unregularized},  UFOs can accelerate grokking by more than $100\times$ when used without weight decay and compared to an unregularized baseline. \citet{lee2024grokfastacceleratedgrokkingamplifying} also show that their method can accelerate grokking by $\sim50.5\times$ when combined with weight decay and compared to an unregularized baseline. Although not explicitly reported in Table~\ref{tab:grokking_treatments} or ~\ref{tab:grokking_unregularized}, the combined speedup caused by weight decay and the selective intracomponent attention UFO is $\sim157\times$ when compared to an unregularized baseline without weight decay, greater than the $\sim50.5 \times$ reported by \citet{lee2024grokfastacceleratedgrokkingamplifying}. \citet{kazanskii2026tamegrokkingrepresentationgeometry}'s method (described in Section~\ref{sec:related-work}) when combined with weight decay can accelerate grokking in transformers by $\sim2 \times$ when compared with a weight-decay-only baseline,\footnote{\citet{kazanskii2026tamegrokkingrepresentationgeometry}'s method can accelerate grokking by larger multiples in residual MLPs, by up to $52\times$ compared to $\sim2\times$ in transformers. The UFO experiments here were tested on transformers only.} while intracomponent attention UFOs can accelerate grokking by $\sim12.3\times$ when combined with weight decay and compared with a weight-decay-only baseline. As such, UFOs compare favorably to other methods designed to accelerate grokking. Moreover, as UFOs may leverage a different mechanism to accelerate grokking, they plausibly complement other methods in addition to weight decay. That said, the aim of the grokking experiments here is not simply to maximize the efficiency of grokking, but rather to provide further evidence for the influence of regular representation on learning, whose implications may extend well beyond grokking.

On the whole, the dramatic grokking speedups induced by UFOs suggest that UFOs are a novel method for improving generalization and the underlying quality of representations.  
\section{Limitations}
\label{sec:limitations}

The UFO method requires additional complexity and computational overhead because the UFO $g_\phi$ must be trained and used to compute gradients to update the primary network $f_\theta$. This additional computational overhead is most pronounced in the intracomponent case because the large input and output dimensionality significantly increases the UFO parameter count compared to the intercomponent case. A particular shortcoming of the intracomponent implementation is that it requires a fixed input position in the representation vector $r_{k, \mathrm{intra}}$ for every possible input in the domain, which is not scalable to arbitrarily large domains. Moreover, the intracomponent case in the modular arithmetic setting has the shortcoming that only some of the inputs to $r_{k,\mathrm{intra-MLP}}$ or $r_{k,\mathrm{intra-attn}}$ are filled with hidden node activations or attention weights, and the zero-filled positions are likely to make  $g_\phi$'s learning more difficult. 

On the whole, in the grokking modular division experiments the intercomponent treatments are more reliable and scalable than their intracomponent counterparts. Nevertheless, the intracomponent approach presented here may be a stepping stone to more scalable approaches to regularizing individual network components. The possibility of regularizing each component to compute a regular function of the network inputs remains intriguing, as it would imply that the neural network that is composed of these regular components would itself compute a regular function. 

Another issue is that the implementations presented here rely on fixed-length sequences wherein there are consistent roles for each position in the sequence across inputs. Overall, UFOs may need to be adapted in the future to work with larger networks and in contexts with more diverse data.

\section{Related Regularization Methods}
\label{sec:related-work}

Related to UFOs, some prior work also regularizes networks through shaping their representations. Some methods  encourage activation sparsity (e.g.\ through L1 penalties on activations) \citep{pmlr-v15-glorot11a, georgiadis2019acceleratingconvolutionalneuralnetworks, pmlr-v119-kurtz20a, song2025prosparseintroducingenhancingintrinsic}; other methods penalize activation norms or the distance between hidden states at consecutive timesteps \citep{Merity2017RevisitingAR} or minimize the correlation between hidden units \citep{cogswell2016reducingoverfittingdeepnetworks}. Unlike these fixed penalties, UFOs aim to learn the representation regularizer itself. 

More closely related to UFOs, another line of work explores regularizers that encourage low-dimensional hidden layer activation geometry \citep{zhu2017ldmnetlowdimensionalmanifold,sanyal2020robustnessdeeplowrankrepresentations,bhalodia2019coopsubnetcooperatingsubnetworkdatadriven, kazanskii2026tamegrokkingrepresentationgeometry}. In particular, prior work by \citet{sanyal2020robustnessdeeplowrankrepresentations} and \citet{bhalodia2019coopsubnetcooperatingsubnetworkdatadriven} and concurrent work by \citet{kazanskii2026tamegrokkingrepresentationgeometry} are similar in approach to intercomponent MLP UFOs. Like UFOs, \citet{sanyal2020robustnessdeeplowrankrepresentations} and \citet{bhalodia2019coopsubnetcooperatingsubnetworkdatadriven} co-train a low-capacity auxiliary model to reconstruct a hidden layer's activations of a primary network and backpropagate the reconstruction error to the primary network, pushing the primary network to have hidden layer activations that the low-capacity auxiliary model can capture. \citet{sanyal2020robustnessdeeplowrankrepresentations}'s auxiliary model is linear, and \citet{kazanskii2026tamegrokkingrepresentationgeometry} similarly pushes each hidden layer's activations toward a low-dimensional linear subspace. Rather than training a capacity-limited auxiliary model like \citet{sanyal2020robustnessdeeplowrankrepresentations}, \citet{kazanskii2026tamegrokkingrepresentationgeometry} penalizes activation variance within a layer that is not captured by top principal components, and notably \citet{kazanskii2026tamegrokkingrepresentationgeometry}'s regularizer also accelerates grokking (see Section~\ref{sec: grokking results}). Because the activations of networks that generalize well tend to lie near curved manifolds \citep{ansuini2019intrinsicdimensiondatarepresentations}, a non-linear autoencoder is likely to better model layer activations than the linear approaches of \citet{sanyal2020robustnessdeeplowrankrepresentations} and \citet{kazanskii2026tamegrokkingrepresentationgeometry}. \citet{bhalodia2019coopsubnetcooperatingsubnetworkdatadriven} use a non-linear autoencoder, and their method most closely resembles the intercomponent MLP UFOs introduced here.

However, UFOs differ from this prior work in key ways, perhaps most centrally in their motivations. UFOs are rooted in the concepts of UFR and FER, and in attempting to promote UFR connect to a broader view of representational quality introduced by \citet{kumar2025questioningrepresentationaloptimismdeep}. While prior methods \citep{zhu2017ldmnetlowdimensionalmanifold,sanyal2020robustnessdeeplowrankrepresentations,bhalodia2019coopsubnetcooperatingsubnetworkdatadriven, kazanskii2026tamegrokkingrepresentationgeometry} are motivated by the hypothesis that regularizers should encourage low-dimensional structure in hidden layer activations, UFOs are motivated by the more general hypothesis that regularizers should encourage regular representations that can be modeled by an auxiliary UFO network. These prior methods only regularize hidden layer activations, but this more general hypothesis leads to novel UFO designs that also regularize attention patterns, individual attention edges, and how \emph{individual} hidden nodes respond across training examples. 

Even intercomponent MLP UFOs, the UFO variant that is most similar to these prior methods, differ in how they are trained and applied. \citet{sanyal2020robustnessdeeplowrankrepresentations} and \citet{bhalodia2019coopsubnetcooperatingsubnetworkdatadriven} train their auxiliary models to reconstruct hidden layer activations from all training examples, and the principal components used by \citet{kazanskii2026tamegrokkingrepresentationgeometry}'s penalty are also determined by all training examples. In contrast, the UFO is trained selectively in an attempt to provide greater regularization pressure on irregular representations, rather than accommodating them (Section~\ref{sec:selective-ufo-updates}). Further, \citet{sanyal2020robustnessdeeplowrankrepresentations} and \citet{bhalodia2019coopsubnetcooperatingsubnetworkdatadriven} apply their regularizer to only one or two hidden layers in their experiments, but the UFO regularizer, like that of \citet{kazanskii2026tamegrokkingrepresentationgeometry}, is applied to all layers. By regularizing every layer, UFOs aim to mitigate FER throughout the network.

Also related is a method by \citet{lai2025saereg}, who co-train a small transformer together with sparse autoencoders (SAEs) \citep{cunningham2023sparseautoencodershighlyinterpretable, bricken2023monosemanticity} placed throughout the network, instead of training SAEs post hoc on an already-trained transformer as is typical. In contrast to UFOs, these SAEs target residual stream activations rather than MLP activations or attention weights. The co-trained SAEs encourage the transformer's residual stream activations to be sparsely decomposable into features of an \emph{overcomplete} SAE, rather than compressible by a low-dimensional bottleneck as in UFOs. Further, the co-trained SAEs are intended to make the model more interpretable and to reduce SAE reconstruction error, not to improve the capabilities of the transformer (although \citet{lai2025saereg} do observe a slight improvement in the transformer's loss as well).

Related to the grokking experiments in this paper, a smaller body of prior work explores methods that directly regularize attention patterns \citep{xu2016showattendtellneural,lin2017structuredselfattentivesentenceembedding,attanasio2022entropybasedattentionregularizationfrees}. Most similar to UFOs that regularize attention, \citet{deshpande2020guidingattentionselfsupervisedlearning} encourage attention heads to follow a pre-determined set of patterns. Each of these prior methods pre-specifies a property of good attention and designs a regularizer to encourage it. In contrast, the regularities in attention encouraged in UFO-based training are not pre-specified but are instead learned by the UFO network.

Additionally, UFOs are related to methods that train a network to model the representations of another network. For example, \citet{ganin2016domainadversarialtrainingneuralnetworks} improve domain transfer by training a network so that a co-trained classifier cannot distinguish which domain the hidden features of that main network came from. Other methods instead train a model to predict its own hidden states or the hidden states of a moving-average copy of itself and observe that
this approach can lead to useful representations
\citep{grill2020bootstraplatentnewapproach, baevski2022data2vecgeneralframeworkselfsupervised, assran2023selfsupervisedlearningimagesjointembedding}, more compact representations \citep{teoh2026nextlatentpredictiontransformerslearn}, or reduced network complexity \citep{premakumar2024unexpectedbenefitsselfmodelingneural}, findings that are consistent with UFOs inducing better representations. 

Similarly, UFOs are related to methods where one network interacts with the weights of another network. For example, \citet{schmidhuberfastweights} introduces a ``fast weights'' method wherein one network updates the weights of another, and in Hypernetworks \citep{ha2017hypernetworks} and its predecessor HyperNEAT \citep{hyperneat} one network encodes the weights of another. In contrast to these methods, UFOs do not directly write or update weights, but instead partially influence the weights by learning to compress representations found in patterns of activations. UFOs can leverage such representations as a form of training data to better influence (but not completely dominate) the primary network. More broadly, activation-level patterns may be a large, potentially underexplored source of training data for improving model capability.

On the whole, representation regularizers that directly influence a network's hidden activations or attention patterns have shown promising results in improving network robustness \citep{sanyal2020robustnessdeeplowrankrepresentations}, sample efficiency \citep{zhu2017ldmnetlowdimensionalmanifold, bhalodia2019coopsubnetcooperatingsubnetworkdatadriven, deshpande2020guidingattentionselfsupervisedlearning}, and generalization  \citep{cogswell2016reducingoverfittingdeepnetworks, kazanskii2026tamegrokkingrepresentationgeometry}, providing grounds for the belief that representation regularization may be helpful in encouraging UFR. 

Nevertheless, representation regularizers are far less common than their weight-based counterparts (such as weight decay). Representation regularizers may be overlooked in part because there is no overarching theory for why they would be uniquely important. By providing such a framework, the concepts of FER and UFR offer a novel lens for recognizing their significance. In this way, this paper aims to connect related methods to FER and UFR and highlight the potential of representation regularization to address fundamental shortcomings of neural network training. 

\section{Discussion and Conclusion}
\label{sec:discussion}

The original paper from \citet{kumar2025questioningrepresentationaloptimismdeep} introducing the \textit{fractured entangled representation hypothesis} (FERH) speculated that the circuitous, open-ended \emph{trajectory} (or \emph{path}) of search in Picbreeder (which is very different from conventional gradient descent) is likely a key part of the explanation for the UFR its representations exhibit.  In short, it could be that the path matters as much as the destination for the purpose of learning good representations.  However, this argument left a profound problem unresolved: what kind of algorithm can induce more Picbreeder-like trajectories through the weight space?  The open-ended nature of search in Picbreeder seems to imply  that conventional gradient descent may even be disqualified from producing such trajectories by being so rigidly objective-driven.

The results with UFOs in this paper are exciting in part because they seem to offer a surprisingly simple solution to this problem that does not require discarding gradient descent.  In effect, UFOs pull the trajectory of the search away from simply climbing the task optimization gradient towards a more elegant representational ideal.  This tug of war between task-driven representation learning and regularity in representation seems to be enough to dramatically impact underlying representation and ultimately yield activation patterns that look more like those in Picbreeder-evolved networks.  The experiments in grokking, which echo related work in representation regularization \citep{zhu2017ldmnetlowdimensionalmanifold, bhalodia2019coopsubnetcooperatingsubnetworkdatadriven, sanyal2020robustnessdeeplowrankrepresentations, deshpande2020guidingattentionselfsupervisedlearning, kazanskii2026tamegrokkingrepresentationgeometry}, confirm that the effect is not only aesthetic and indeed has significant downstream implications for learning and generalization.  We expect further implications to be uncovered in the future regarding continual learning and creativity.  

UFOs are motivated by the hypothesis that regular activation and attention patterns are essential to good representation. The FER evident in the conventional gradient descent baseline representations shown in Figure \ref{fig:sgd-butterfly} suggests that we should not expect UFR to emerge from conventional training. Consequently, a natural question is the extent to which FER is present in frontier LLMs and whether it is at least to some extent corrected by scaling data and compute. 
One clue from Goodfire, Anthropic, and others is the identification of structured activation manifolds corresponding to LLM representations of counting, days of the week, color, and more \citep{gurnee2025when, bhalla2026sparseautoencoderscaptureconcept, modell2025originsrepresentationmanifoldslarge, engels2025not}. Consistent with the hypothesis that regular activations are essential to good representation, these low-dimensional activation manifolds suggest that LLMs have learned good representations of these topics.

However, despite these findings of regular activation patterns and the increasingly impressive performance of LLMs, at the same time LLMs continue to show evidence of FER. For example, \citet{dong2026revisitingreliabilitylanguagemodels} show that GPT-5's instruction-following performance drops by 18.3\% when non-semantic changes are made to a prompt. More striking evidence of FER is prevalent in older or smaller models. GPT-3's performance on simple counting tasks could swing wildly depending on the type of object that was being counted, suggesting that the model learned disparate, object-dependent heuristics for counting, rather than a unified representation of counting \citep{kumar2025questioningrepresentationaloptimismdeep}. As tasks that previously showed evidence of FER inevitably enter the training pipelines for the next generation of LLMs, evidence for FER becomes more difficult to identify. Modern LLMs can also obfuscate underlying FER by achieving broad competency through combining a large number of jumbled heuristics, rather than learning general representations. Better measures of FER are needed and must continue to evolve to keep pace with rapid LLM development. 

One possible explanation of the contradictory evidence for FER or UFR in LLMs could be the amount of training data available for specific topics.  It is possible that well-covered topics more closely approach UFR while sparsely covered topics tend towards FER. Consistent with this hypothesis,
\citet{power2022grokkinggeneralizationoverfittingsmall} observe that the number of training steps needed to grok increases exponentially as the training data size decreases. Further, work by \citet{morris2025languagemodelsmemorize} argues that language models memorize their training data until dataset size passes a threshold relative to model size, after which a grokking-like transition to generalization begins. In domains where there is less training data available, a much larger set of functions is likely to fit the data, and many of those functions are likely to exhibit FER. 

\begin{figure*}[!htbp]
    \centering
    \includegraphics{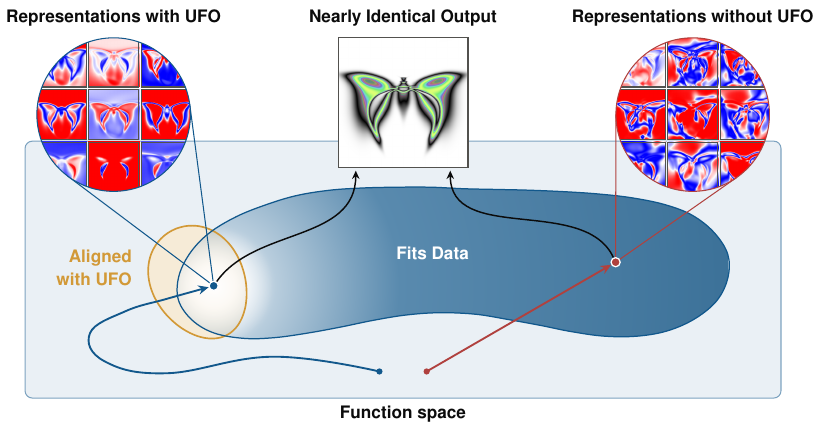}
    \captionsetup{skip=4pt}
    \caption{\textbf{UFOs help navigate to parts of function space that fit the data with regular representations}. The light blue rectangle depicts all of function space, and the darker blue shape depicts the functions that fit the training data. In some sense, conventional loss functions are \emph{underspecified} --- they will guide training towards a function that fits the data, but do not additionally specify that the function that fits the data should also be regular. When a loss function is augmented with a UFO, it can better guide search to a location in function space that both fits the data and has regular representations.}
    \label{fig:function-space}
\end{figure*}

While the implication of this analysis seems to be that representation is at the mercy of the quantity and quality of data available for a given topic, UFOs potentially offer a way out.  In effect, typical loss functions (e.g.\ cross entropy or mean-squared error) are \emph{underspecified}. That is, these loss functions ask that the network fit the training data, but do not additionally request that the network fit the training data with regular representations. As depicted in Figure~\ref{fig:function-space}, the UFO method aims to address this shortcoming by providing an additional, learned constraint on the representations of the network fitting the training data. If FER is more likely in low-data regimes, UFOs may be particularly valuable in such settings, a finding that would be consistent with results of the representation regularizers introduced by 
\citet{zhu2017ldmnetlowdimensionalmanifold}, \citet{bhalodia2019coopsubnetcooperatingsubnetworkdatadriven}, 
and \citet{deshpande2020guidingattentionselfsupervisedlearning}. If UFOs improve underlying representation without the need for so much data, then UFOs could significantly improve the sample efficiency of neural networks, a potentially exciting area for future investigation. Furthermore, techniques that improve the sample efficiency of LLMs may be crucial for enabling greater LLM creativity. Because data is inevitably limited at the frontier of knowledge, creatively pushing forward that frontier requires forming good representations without the crutch of big data. To be truly creative, models must form representations with regularities that extend beyond the limited data that has been observed, capturing the world as it could be, rather than merely how it has been.  

Although in this paper UFOs are only applied to small models in small domains, UFOs may be powerful in regularizing large models. Intuitively, if two models both achieve the same training loss, one small and one large, the small model is more likely to generalize well because it has fewer parameters for overfitting. However, in practice, it is difficult to train small models to achieve the same low loss as large models; the small model's optimization landscape is much less forgiving, and even if optimization finds the global minimum, the small model may lack the capacity to match the large model's loss. Perhaps what we want then is the easy optimization of a large model and the compressed representations of a small model, where possible. In asking that a model compress its representation as much as possible while achieving low loss, UFOs may be able to achieve this desirable combination of optimization ease and representation quality. An important area of future work is to focus on scaling up and adapting UFOs to larger models and domains, including LLMs. 

Even though learning good representations is clearly critical for AI, the mechanistic underpinnings of FER and UFR remain underinvestigated.  The field of mechanistic interpretability has made notable progress in better understanding neural net representations \citep{10136140, sharkey2025openproblemsmechanisticinterpretability}, but mechanistic interpretability focuses to a large extent on \emph{describing} representations, without passing a value judgment on whether a representation is high or low quality. For example, in contrast to UFO autoencoders that are co-trained with another network to encourage better representations in that network, SAEs are commonly trained post hoc to describe the internal activations of an already-trained network \citep{cunningham2023sparseautoencodershighlyinterpretable, bricken2023monosemanticity}.

When attempting to describe representations, researchers may implicitly adopt an attitude of representational optimism \citep{kumar2025questioningrepresentationaloptimismdeep}, assuming that the network under study has learned good representations. Such representational optimism may lead researchers to conclude that uninterpretable representations suggest a shortcoming in their interpretability tools, rather than a shortcoming of the network's representations, inferring that the network has learned alien or opaque representations but not low-quality representations. And yet, if a network has learned FER, then of course it will be difficult to neatly model the network's representations because the network has only learned a jumble of heuristics. That is \textit{not} to say that uninterpretable representations are bad representations, or that the characteristics that make representations difficult to interpret (e.g.\ polysemanticity; \citealp{Olah2020ZoomIA}) are correlated with FER. Rather, an insufficient mechanistic understanding of what is essential to good representation prevents us from properly adjudicating between the case where the network has learned FER and the case where the network has learned alien but nevertheless high-quality representations. 

Furthermore, if we assume that deep learning inevitably leads to high-quality representations, we rob ourselves of the opportunity to develop a mechanistic understanding of the shortcomings of modern neural networks. Of course, neural networks can achieve extraordinary performance, but human-level sample efficiency, continual learning, and creativity are still presently challenging. The field of mechanistic interpretability should be \textit{more ambitious} and aim to develop a normative theory of representation that builds on the descriptive theory that the field is currently developing. With such a theory, the field would go beyond describing representations as they are and evaluate how they should be. Developing this mechanistic theory of UFR and FER may be an important step in correcting FER and ultimately making deep learning even more powerful.  

Interestingly, the UFO's results suggest that promoting regular representations that are learnable by another capacity-limited network can be a powerful lever in improving representations and accelerating generalization. Under certain conditions, interpretable representations may be a signature of UFR, presenting the intriguing possibility that a model could be trained to be more capable \textit{because} it was trained to be more interpretable by a UFO. Prior work has explored methods that train a network to be more interpretable but often implies that interpretability is orthogonal to capability \citep{pmlr-v235-tamkin24a, jermyn2022engineeringmonosemanticitytoymodels, chen2025lowrankadaptingmodelssparse, lai2025saereg}. \citet{chen2025lowrankadaptingmodelssparse} and \citet{lai2025saereg} train a language model to be interpretable by an SAE, but both methods aim to improve the reconstruction fidelity of the SAE, not the capabilities of the language model. UFO designs that make models both more capable and more interpretable may be an exciting direction for future work. 

To intuitively illustrate why interpretable representation may be a signature of more capable representation in some settings, consider the following analogy. Suppose there are two students, both of whom receive the same score on their math exam. While the first student can clearly explain her solution to every question on the exam to her classmates, the second student is completely incapable of explaining his solutions to anyone (even with the painstaking effort of his classmates over many conversations). Although their exam scores alone cannot distinguish their understanding, the rest of the class will likely conclude that the first student has learned more powerful representations because her representations are more interpretable to them. Even if the second student has learned equally powerful representations, his representations may be improved through the exercise of clear explanation. Indeed, explaining a subject clearly to others is an effective way to learn that subject better \citep{Bargh1980OnTC, FIORELLA2013281}. Analogously, neural net representations that are more easily modeled by an auxiliary UFO network may be evidence of more capable representations, and the pressure to make these representations more interpretable may further improve learning. 

Perhaps the most intriguing question is
what it would imply for the capabilities of LLMs if they could represent their pretraining data with the same simplicity and regularity as seen in the original Picbreeder network representing the butterfly, as shown in Figure~\ref{fig:picbreeder-butterfly}. If the FERH is correct and LLMs are still riddled with FER, then even frontier LLM capabilities could potentially be dramatically advanced by promoting UFR. How well networks generalize, learn, and imagine is ultimately a function of their representations. We believe that there is significant room remaining to improve UFOs in the future and thereby train loss functions that promote UFR. Our hope is that this work will serve as a stepping stone to better UFO designs than those presented here, and to further research that explores the implications of UFO-induced representations. 

\clearpage
\bibliographystyle{iclr2026_conference}
\bibliography{references}

\clearpage
\onecolumn

\appendix
\section{Normalizing Constants for UFO Losses}
\label{app:normalizers}
\vspace{-0.5em}
Both $\mathcal{L}_\mathrm{aux}$ and $\mathcal{L}_\mathrm{select}$ are mean-squared errors, so both $\mathcal{L}_\mathrm{aux}$ and $\mathcal{L}_\mathrm{select}$ are normalized by the number of squared errors they sum. In the intracomponent case, there are $K$ representations, one per hidden component, and each has $n$ entries, one per training example in $\mathcal{B}$. $\mathcal{L}_\mathrm{aux}$ therefore sums $Kn$ squared errors, and $\mathcal{L}_\mathrm{select}$ sums $K n_{\mathrm{sel}}$ squared errors because $n_{\mathrm{sel}}$ entries are selected in each representation.

In the intercomponent case, let $L$ be the number of layers in $f_\theta$ and $c$ the number of chosen components in each layer. Each layer contributes one representation per training example, so there are $K = nL$ representations, each with $c$ entries. $\mathcal{L}_\mathrm{aux}$ therefore sums $Kc$ squared errors and is normalized by $Kc$. Because $n_{\mathrm{sel}}$ representations are selected in each of the $L$ layers, $\mathcal{L}_\mathrm{select}$ sums $L\, n_{\mathrm{sel}}\, c$ squared errors and is normalized by $L\, n_{\mathrm{sel}}\, c$.

\section{Additional Representation Visualizations From the Picbreeder Image Regression Domain}
\label{app:picbreeder}
\vspace{-0.5em}

This appendix provides additional representation visualizations from the Picbreeder image regression domain and discussion to complement the Picbreeder experiments presented in Section~\ref{sec:picbreeder-experiments}. 

\begin{figure}[p]
    \centering
    \captionsetup{skip=3pt}
    \captionsetup[subfigure]{skip=2pt}
    \begin{subfigure}{\textwidth}
        \centering
        \includegraphics[
            width=\textwidth,
            height=\featuremapheight,
            keepaspectratio
        ]{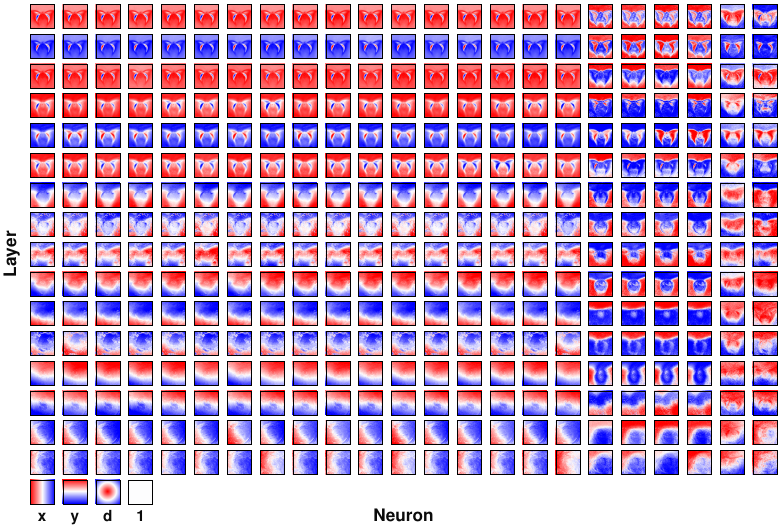}
        \caption{Intracomponent}
        \label{fig:reconstructed-intracomponent}
    \end{subfigure}
    \par\vspace{0.3em}
    \begin{subfigure}{\textwidth}
        \centering
        \includegraphics[
            width=\textwidth,
            height=\featuremapheight,
            keepaspectratio
        ]{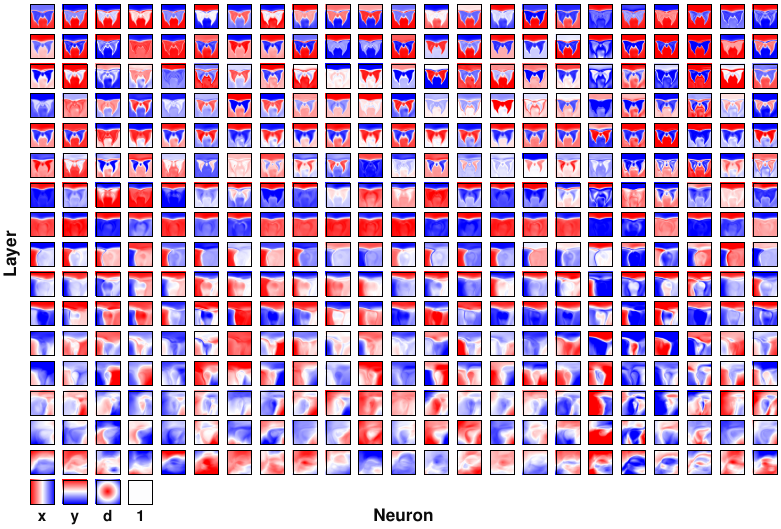}
        \caption{Intercomponent}
        \label{fig:reconstructed-intercomponent}
    \end{subfigure}
    \par\vspace{0.2em}
    \caption{\textbf{Representations reconstructed by UFO autoencoders.} These representation visualizations depict reconstructed activations from the intracomponent (a) and intercomponent (b) UFO autoencoders in Section~\ref{sec:picbreeder-experiments}. As can be seen, the UFO autoencoders do reasonably well at reconstructing the activations of the networks they regularize, and these reconstructed activations can be compared to the actual activations shown in Figures~\ref{fig:intracomponent-ufo-butterfly} and~\ref{fig:intercomponent-ufo-butterfly}.
    }
    \label{fig:ufo-reconstructed-representations}
\end{figure}

To illustrate the fidelity with which the UFO autoencoder $g_\phi$ can reconstruct the primary network $f_\theta$'s activations, Figure~\ref{fig:ufo-reconstructed-representations} visualizes the activations that are reconstructed by the autoencoder $g_\phi$ for the intracomponent and intercomponent regularized CPPNs featured in Section~\ref{sec:picbreeder-experiments}. These reconstructed activations can be compared to the actual activations visualized in Figure~\ref{fig:ufo-butterfly-feature-maps}. As can be seen, despite aggressive bottlenecks (a $2$-dimensional bottleneck in the intracomponent case and a $4$-dimensional bottleneck in the intercomponent case), the UFO $g_\phi$ does reasonably well in reconstructing $f_\theta$'s activations, although in the intracomponent case, reconstructed hidden node activation maps are fuzzier and less well-defined than in the intercomponent case. 

\begin{figure}[p]
    \centering
    \captionsetup{skip=3pt}
    \captionsetup[subfigure]{skip=2pt}
    \begin{subfigure}{\textwidth}
        \centering
        \includegraphics[
            width=\textwidth,
            height=\featuremapheight,
            keepaspectratio
        ]{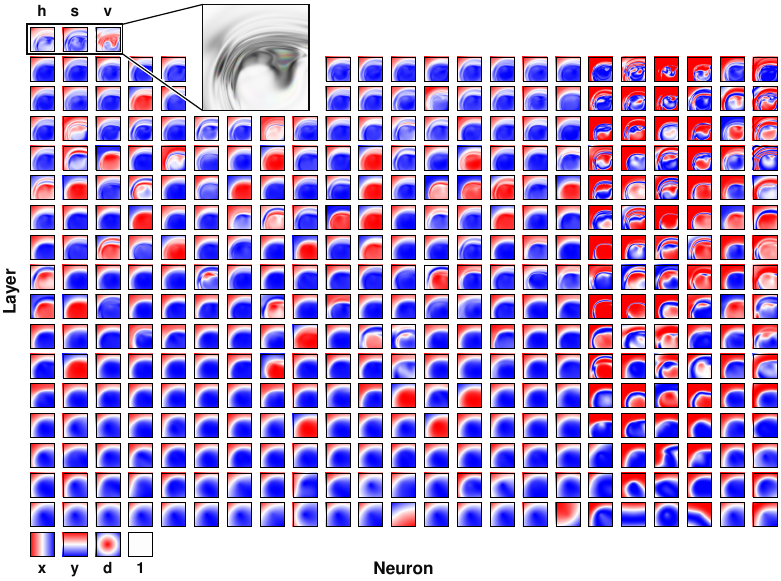}
        \caption{Intracomponent}
        \label{fig:underfit-intracomponent-ufo-butterfly}
    \end{subfigure}
    \par\vspace{0.3em}
    \begin{subfigure}{\textwidth}
        \centering
        \includegraphics[
            width=\textwidth,
            height=\featuremapheight,
            keepaspectratio
        ]{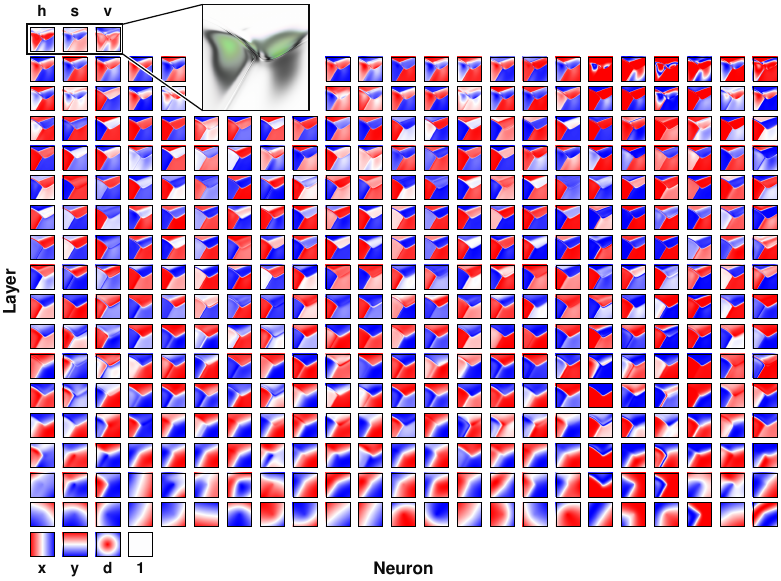}
        \caption{Intercomponent}
        \label{fig:underfit-intercomponent}
    \end{subfigure}
    \par\vspace{0.2em}
    \caption{\textbf{UFO-regularized networks that underfit the target butterfly.} These visualizations show the output and representations of intracomponent (a) and intercomponent (b) UFO-regularized networks that underfit the target butterfly. If hyperparameters are not tuned appropriately or there is an unlucky seed, UFO regularization can lead to considerable underfitting or subpar representations. These networks use the same layerized architecture and visualization scheme as the original Picbreeder butterfly network shown in Figure~\ref{fig:picbreeder-butterfly}.}
    \label{fig:underfit-ufo-regularized-networks}
\end{figure}

As mentioned in Section~\ref{sec:picbreeder-experiments}, the representational quality of UFO-regularized networks can vary depending on the random seed used for network initialization, and if hyperparameters are not tuned appropriately the UFO can lead to considerable underfitting. Figure~\ref{fig:underfit-ufo-regularized-networks} shows examples where UFO regularization led to such underfitting. 

\begin{figure}[p]
    \centering

    \begin{subfigure}{\textwidth}
        \centering
        \includegraphics[
            width=\textwidth,
            height=\featuremapheight,
            keepaspectratio
        ]{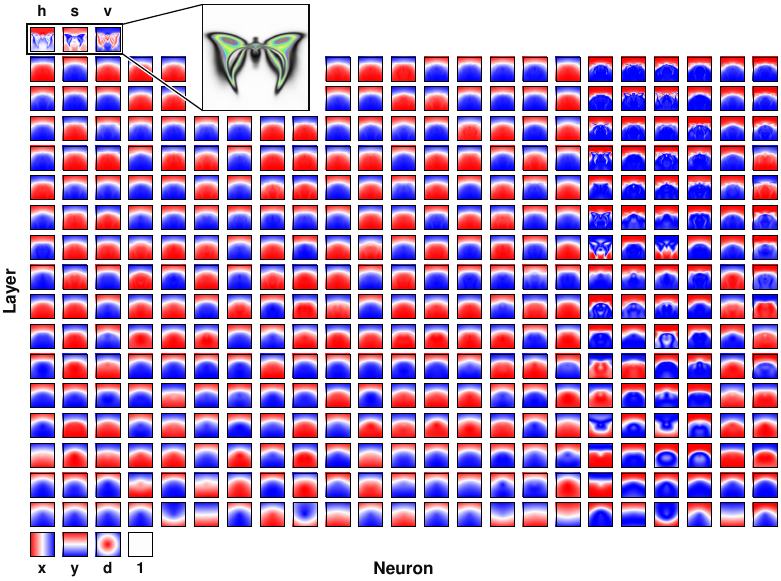}
        \caption{Intracomponent}
        \label{fig:alternative-intracomponent-ufo-butterfly}
    \end{subfigure}   

    \vspace{0.8em}

    \begin{subfigure}{\textwidth}
        \centering
        \includegraphics[
            width=\textwidth,
            height=\featuremapheight,
            keepaspectratio
        ]{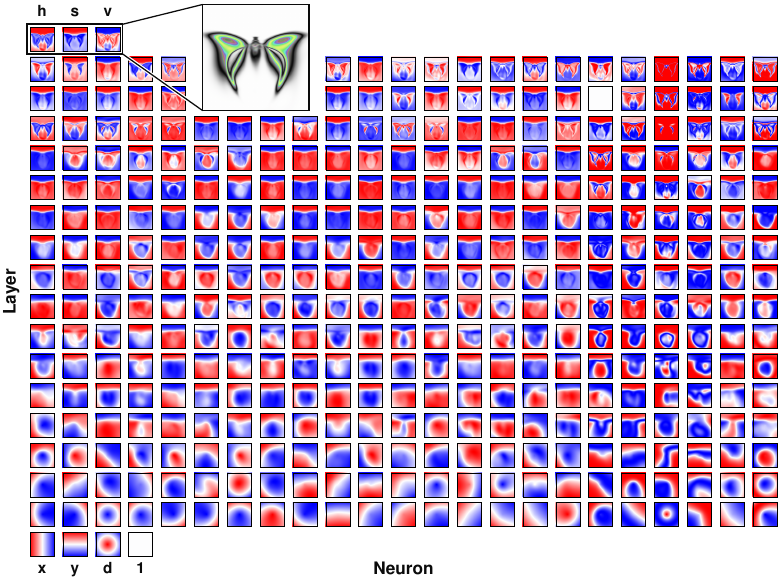}
        \caption{Intercomponent}
        \label{fig:alternative-ufo-reglurization-feature-maps}
    \end{subfigure}

    \caption{\textbf{Representations of additional UFO-regularized networks trained to output the butterfly.} Notice that the intracomponent UFO-regularized network (a) features many redundant hidden node activation maps, and the butterfly's main shape is constructed by a small number of hidden nodes; the intercomponent UFO-regularized network (b) has many hidden nodes that elaborate different parts of the butterfly's structure. These networks use the same layerized architecture and visualization scheme as the original Picbreeder butterfly network shown in Figure~\ref{fig:picbreeder-butterfly}.} 
    \label{fig:alt-ufo-regularization-feature-maps}
\end{figure}

The representations shown in Figure~\ref{fig:ufo-butterfly-feature-maps} visualize the results of UFO regularization. Different kinds of regular representations can also be obtained through UFO regularization, and Figure~\ref{fig:alt-ufo-regularization-feature-maps} provides additional representation visualizations from UFO-regularized networks. Networks with different representations will have different affordances, and UFOs provide a mechanism to explore the space of representations of a given training dataset. An exciting area of future work is to better understand the implications of different UFO-induced representations, which in turn could help develop better UFO designs. 

\begin{figure}[!t]
    \centering
    \includegraphics[width=\textwidth]{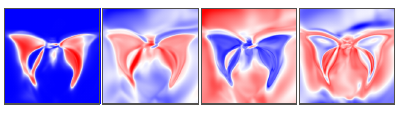}
    \caption{\textbf{Regular hidden node activation maps from the conventional gradient
    descent trained network in Figure~\ref{fig:sgd-butterfly}.}
    Although this paper emphasizes the irregular representations learned by conventional gradient descent, select hidden nodes from the conventional gradient descent trained network shown in Figure~\ref{fig:sgd-butterfly} do capture some of the symmetry of the butterfly. That conventional gradient descent leads to \textit{some} representational regularity is helpful for the UFO because it may be able to identify and amplify these regularities.
    \label{fig:regular-conventional-gradient-descient-hidden-nodes}
    }
\end{figure}

Although this paper emphasizes the irregular representations learned by conventional gradient descent, select hidden nodes from the conventional gradient descent trained network shown in Figure~\ref{fig:sgd-butterfly} do capture some of the symmetries of the butterfly.
Some of these more regular hidden nodes are shown in Figure~\ref{fig:regular-conventional-gradient-descient-hidden-nodes}. That conventional gradient descent leads to \textit{some} representational regularity is promising because the UFO may be able to identify and amplify these regularities.


\section{Additional Grokking Division Modulo 97 Training Details}
\label{app:grokking}
\vspace{-0.5em}

This appendix provides additional training details for the grokking modular division experiments presented in Section~\ref{sec:grokking-experiments}. 

The primary network $f_\theta$ is a 2-layer, decoder-only transformer with a model embedding dimension of 128, MLP blocks with 512 hidden nodes and ReLU activations, attention blocks with four attention heads, and layer norm \citep{ba2016layernormalization} after each attention and MLP block. Token embeddings are learnable parameters, and the positional embeddings are fixed and sinusoidal. Minibatches of $512$ problems per step train $f_\theta$ and AdamW is the optimizer with Adam betas of (0.90, 0.98) and an Adam epsilon of $10^{-8}$. The learning rate is $10^{-3}$ and it is raised linearly over 10 warm-up steps. Experiments with weight decay use a weight decay of 0.03 and the decay is ``to zero'' rather than ``to init,'' and is applied to all learnable parameters. 

Like the transformer $f_\theta$, the UFO autoencoder $g_\phi$ is trained with a learning rate of $10^{-3}$, but without a learning rate warmup. The UFO $g_\phi$ is trained with the Adam optimizer with PyTorch's default Adam betas of (0.90, 0.999) and does not use weight decay or other regularization. 

As described in Section~\ref{sec:grokking-experimental-setup}, there are tokens for each element of $\mathbb{Z}_{97}$, for the \texttt{=} and \texttt{/} symbols, and a special end-of-sequence token \texttt{<|eos|>}. The division modulo 97 dataset consists of token sequences of the form \texttt{<|eos|> x / y = z <|eos|>}. The transformer $f_\theta$ is tasked with minimizing the cross entropy loss over the correct next token \texttt{z} when given the prefix \texttt{<|eos|> x / y =} and predicting \texttt{<|eos|>} when given the prefix \texttt{<|eos|> x / y = z}. In calculating train and validation accuracy, a problem is counted as correct only if both the \texttt{z} token and \texttt{<|eos|>} token are predicted correctly. 

There are a total of $97\cdot96 = 9$,$312$ problems of division modulo $97$, and $f_\theta$ is trained on half ($4$,$656$) of these problems and evaluated on the other half. A fixed train-val data split is used across all runs, and for seed-matched comparisons the same eight seeds that control the transformer init, UFO init, and minibatch data shuffling are used for each treatment. To include each problem in the training set the same number of times per epoch, a partial batch of $48$ problems is used at the end of each epoch. In the intracomponent case, this partial batch is skipped for UFO regularization and updating, as a representation vector $r_{k,\mathrm{intra}}$ would only have $48$ positions filled in, and such a sparse representation vector may disrupt learning. Because each epoch has nine full batches followed by this partial batch, the intracomponent selectivity schedule spans $30$,$000$ UFO updates over the first $33$,$333$ transformer training steps.

Section~\ref{sec:method} explains that the columns of the component response matrix $R_\theta^{(\ell)}(\mathcal{B})$ are $z$-score normalized. Specifically, $R_\theta^{(\ell)}(\mathcal{B})$ is normalized over a batch $\mathcal{B}$ by taking each column $j$ of $R_\theta^{(\ell)}(\mathcal{B})$, subtracting its column mean $\mu_j^{(\ell)}$, and dividing by its column standard deviation $\sigma_j^{(\ell)}$, yielding a normalized component response matrix. In our implementation, each column $j$ of $R_\theta^{(\ell)}(\mathcal{B})$ is divided by the maximum of the column standard deviation $\sigma_j^{(\ell)}$ and a small $\epsilon = 10^{-6}$ for numerical stability.

To facilitate describing the responses of individual network components in the intracomponent case, every division problem modulo $97$ is assigned a fixed position between $0$ and $9$,$311$ based on the expression $h(x,y) = 96x + (y-1)$, where $x \in \mathbb{Z}_{97}$ and $y \in \mathbb{Z}_{97}\setminus\{0\}$. A component's normalized hidden node activation or attention weight for each problem in a batch $\mathcal{B}$ is assigned to its corresponding position in a $9$,$312$-dimensional vector, where positions corresponding to problems that did not appear in $\mathcal{B}$ are filled with $0$s. This process yields $r_{k, \mathrm{intra-MLP}}$ or $r_{k, \mathrm{intra-attn}}$, respectively. 

In the intercomponent MLP case, a representation vector $r_{k,\mathrm{inter-MLP}}$ is a 512-dimensional vector capturing a layer's normalized MLP activations at the \texttt{=} position, and an intercomponent attention representation vector $r_{k,\mathrm{inter-attn}}$ is 20-dimensional. Under the causal mask, the \texttt{=} query in \texttt{<|eos|> x / y =} attends to 5 positions, including itself, so the \texttt{=}-query row of each attention head is 5-dimensional, and these rows are concatenated across the 4 heads in a layer to form $r_{k,\mathrm{inter-attn}}$. As with MLP activations, attention weights are extracted only at the \texttt{=} position, where $f_\theta$ predicts \texttt{z}, so only the \texttt{=}-query row is regularized. In the second layer of $f_\theta$, the \texttt{=}-query row contains every attention edge that can influence the prediction of \texttt{z}. In the first layer, the attention edges of the earlier queries (\texttt{<|eos|>}, \texttt{x}, \texttt{/}, and \texttt{y}), like the MLP activations at those positions, also influence the prediction of \texttt{z}, but only indirectly, through the second layer's \texttt{=} query reading from those positions. Regularizing these earlier attention edges as well would give the two layers different numbers of attention edges ($15$ per head in the first layer vs.\ $5$ per head in the second layer) and so different dimensions for $r_{k,\mathrm{inter-attn}}$, which would complicate the design of the intercomponent UFO autoencoder $g_\phi$ used for both layers. In the intracomponent case, each attention edge yields its own $9$,$312$-dimensional $r_{k,\mathrm{intra-attn}}$, so a single UFO autoencoder $g_\phi$ could regularize every edge that influences \texttt{z}, but for consistency with the intercomponent case, we still regularize only the \texttt{=}-query row.

Note that in all selective UFO treatments in the grokking experiments, the selectivity schedule $s(t)$ decreases over time, rather than remaining constant as in the Picbreeder intracomponent case described in Section~\ref{sec:picbreeder-experimental-setup}. Also observe that in the intracomponent case $s(t)$ starts off less selective and decreases more slowly than in the intercomponent case. Empirically, this selectivity schedule produced better results for the intracomponent case, suggesting that intracomponent regularities may emerge more slowly or take longer for $g_\phi$ to model than intercomponent regularities in this setting. 

Our implementation of the UFO method in the modular division domain builds on the original grokking code from \citet{power2022grokkinggeneralizationoverfittingsmall}, which was used to train small transformers on a range of algorithmically generated datasets. For ease of extensibility to other domains, our publicly available code (\coderelease) also includes tokens for other algorithmically generated datasets from \citet{power2022grokkinggeneralizationoverfittingsmall}.

\section{UFO Reconstruction Error During Grokking}
\label{app:ufo-aux-trace}
\begin{figure}[!htbp]
    \centering
    \includegraphics{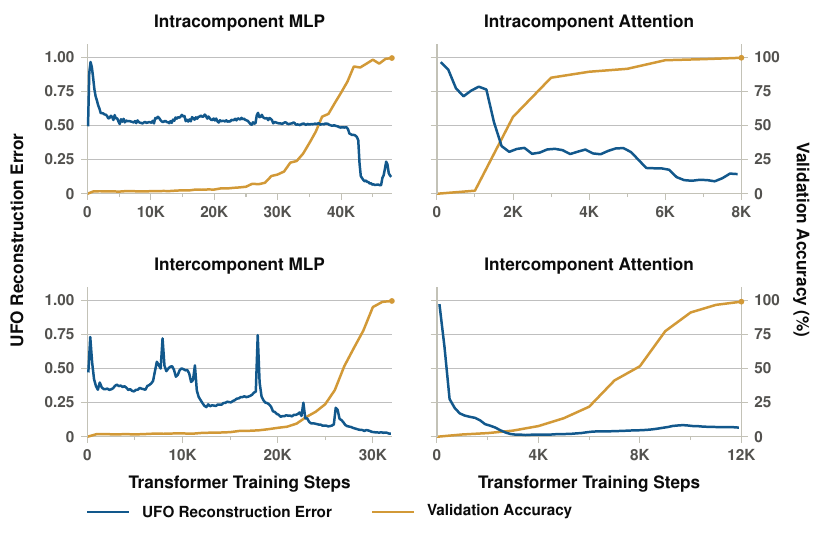}
    \caption{\textbf{UFO reconstruction error falls over training.}  Each panel shows one selective UFO treatment with weight decay, using the seed with the median grok step, until the transformer groks. Blue lines plot the UFO reconstruction error $\mathcal{L}_\mathrm{aux}$ on all of the transformer's representations, averaged over 200-step windows (left axis). Gold lines plot validation accuracy (right axis).}
    \label{fig:ufo-aux-trace}
\end{figure}

\vspace{-0.5em}
To illustrate that the transformer's representations become more compressible by the UFO over training, Figure~\ref{fig:ufo-aux-trace} plots $\mathcal{L}_\mathrm{aux}$ for each selective UFO treatment with weight decay, using the seed with the median grok step, alongside validation accuracy so that changes in $\mathcal{L}_\mathrm{aux}$ can be compared with grokking progress. In all four treatments, $\mathcal{L}_\mathrm{aux}$ decreases substantially over training, ending far below its starting value by the transformer's grok step.

\end{document}